\documentclass[10pt]{article} 
\usepackage[preprint]{tmlr}

\usepackage{amsmath,amsfonts,bm}

\def\eqref#1{equation~\ref{#1}}

\def\va{{\bm{a}}}
\def\vb{{\bm{b}}}

\def\ve{{\bm{e}}}

\def\vm{{\bm{m}}}

\def\vu{{\bm{u}}}
\def\vv{{\bm{v}}}
\def\vw{{\bm{w}}}
\def\vx{{\bm{x}}}
\def\vy{{\bm{y}}}
\def\vz{{\bm{z}}}

\def\evu{{u}}

\def\evx{{x}}

\def\evz{{z}}

\def\mA{{\bm{A}}}

\def\mG{{\bm{G}}}

\def\mJ{{\bm{J}}}
\def\mK{{\bm{K}}}

\def\mM{{\bm{M}}}

\def\mQ{{\bm{Q}}}

\def\mV{{\bm{V}}}
\def\mW{{\bm{W}}}
\def\mX{{\bm{X}}}
\def\mY{{\bm{Y}}}

\DeclareMathAlphabet{\mathsfit}{\encodingdefault}{\sfdefault}{m}{sl}
\SetMathAlphabet{\mathsfit}{bold}{\encodingdefault}{\sfdefault}{bx}{n}

\def\sR{{\mathbb{R}}}

\def\sX{{\mathbb{X}}}

\newcommand{\R}{\mathbb{R}}

\usepackage{hyperref}       
\usepackage{url}            
\usepackage{booktabs}       
\usepackage[dvipsnames]{xcolor}

\usepackage{algorithm}
\usepackage{algorithmicx}
\usepackage{algcompatible}
\usepackage{amsmath}
\usepackage{arydshln}

\usepackage{float}
\usepackage{placeins}

\usepackage{graphicx}

\title{Pulling Back the Curtain on Deep Networks}

\author{\name Maciej Satkiewicz \email maciej.satkiewicz@314.foundation\\
    \addr 314 Foundation
    \AND
    \name Roberto Corizzo\\
    \addr American University
    \AND
    \name Marcin Pietroń\\
    \addr AGH University of Krakow
}

\def\month{MM}  
\def\year{YYYY} 
\def\openreview{\url{https://openreview.net/forum?id=XXXX}} 

\newcommand{\1}[1]{\mathbf{1}_{\{#1\}}}

\usepackage{contour}
\contourlength{0.1pt}
\newcommand{\best}[1]{\contour{black}{#1}}

\newcommand{\pairedci}[3]{%
  \ensuremath{\begin{gathered}
    #1\\[-2pt]
    {\scriptstyle[#2,\;#3]}
  \end{gathered}}%
}
\newcommand{\inconclusive}[1]{\textcolor{blue}{#1}}
\newcommand{\deterioration}[1]{\textcolor{red}{#1}}

\newcommand{\pairsep}{%
  \noalign{\vskip 2pt}%
  \hdashline
  \noalign{\vskip 3pt}%
}

\makeatletter
\newcommand{\apptocsection}[1]{%
  \@dottedtocline{1}{0em}{2.5em}{%
    \hyperref[#1]{%
      \numberline{\ref*{#1}}\textbf{\nameref*{#1}}%
    }%
  }{\hyperref[#1]{\pageref*{#1}}}%
}

\newcommand{\apptocsubsection}[1]{%
  \@dottedtocline{2}{1.5em}{3em}{%
    \hyperref[#1]{%
      \numberline{\ref*{#1}}\nameref*{#1}%
    }%
  }{\hyperref[#1]{\pageref*{#1}}}%
}
\makeatother

\begin{document}

\maketitle

\begin{abstract}
In linear models, visualizing a weight vector naturally reveals the model's preferred input direction, but extending this intuition to deep networks via gradients or GradientAscent often yields brittle or adversarial-looking features.
We argue that deep networks are better understood as input-conditioned affine operators, whose natural adjoint action pulls a neuron's preferred direction back to input space.
We further refine this representation through optional backward-only softening for hard-gated layers and propose pullbacks as an alternative backward operator for existing explanation procedures, including local attribution, smoothing, and iterative ascent.
Across convolutional architectures (ResNet50, VGG) and transformer-based models (PVT), replacing gradients with pullbacks yields improvements in 46 out of 60 comparisons across five faithfulness, stability, and localisation metrics.
In particular, PullbackAscent improves mean scores over GradientAscent across all five primary metrics on each architecture, reducing mean Infidelity by factors of 2.8 to 5.5.
Strikingly, it also produces sharp, perceptually aligned, and target-specific perturbations in only a few steps, without post-processing or explicit perceptual priors.
More broadly, pullback-based methods achieve the best mean in 13 out of 15 architecture-metric combinations, outperforming established attribution baselines.
Our results suggest that adjoint transport can serve as a reusable backward primitive, enhancing existing explanation procedures without architectural changes or elaborate attribution pipelines.
We use \emph{Semantic Pullbacks} (SP) to denote the resulting family of pullback-based explanations. Our code is available at: 
\url{https://github.com/314-Foundation/SemanticPullbacks}
\end{abstract}

\section{Introduction}

Despite remarkable progress in deep learning, understanding what drives the
predictions of modern neural networks remains an open challenge. A dominant
paradigm in post-hoc explainability associates an output score with an
input-space direction, typically its gradient
\citep{simonyan2014deepinsideconvolutionalnetworks}. However, gradients of deep
networks are often noisy, unstable, and highly sensitive to small input
perturbations~\citep{smilkov2017smoothgradremovingnoiseadding}, limiting their
usefulness both as attribution maps and as directions for feature
visualization.

Many explanation methods address these limitations through smoothing,
integration or additional
regularization
\citep{ribeiro2016whyitrustyou,
sundararajan2017axiomaticattributiondeepnetworks,
lundberg2017unifiedapproachinterpretingmodel,
Selvaraju_2019,
shrikumar2019learningimportantfeaturespropagating}.
While effective, such additions make it more difficult to relate the resulting attributions to the network computation. In this work we ask a more basic question: \emph{is the gradient itself the appropriate explanatory primitive?}

For a linear model, the gradient of an output is exactly its weight vector.
A deep network, however, can be viewed locally as an input-conditioned affine
operator with effective linear part $\mW(\vx)$. The full gradient additionally
captures how this operator changes locally with the input. We argue that, when the goal
is to generalize the direct weight-vector interpretation of a linear model, a
more natural object is the adjoint action of the operator currently applied by
the network, i.e. $\nu_\vu(\vx)=\mW(\vx)^\top \vu$, where $\vu$ is an output-space direction. We refer to this transported direction
as the \emph{pullback} of $\vu$.

For piecewise-linear hard-gated operators such as ReLU and MaxPool, the exact
pullback coincides with the standard gradient. We therefore introduce
backward-only softening of these hard routing decisions to obtain a distinct,
more stable explanatory field. This leaves the forward computation and the
classifier's decision boundary unchanged, while allowing weak but consistent
contributions to remain visible. We motivate this softening in detail in
Section~\ref{sec:soft_adjoint} and
Appendix~\ref{app:appendix_motivation}. The resulting layer-wise rules are
closed-form and require no architectural modification, retraining, or
post-processing.

\begin{figure}[tb]
\centering
\includegraphics[width=0.49\linewidth]{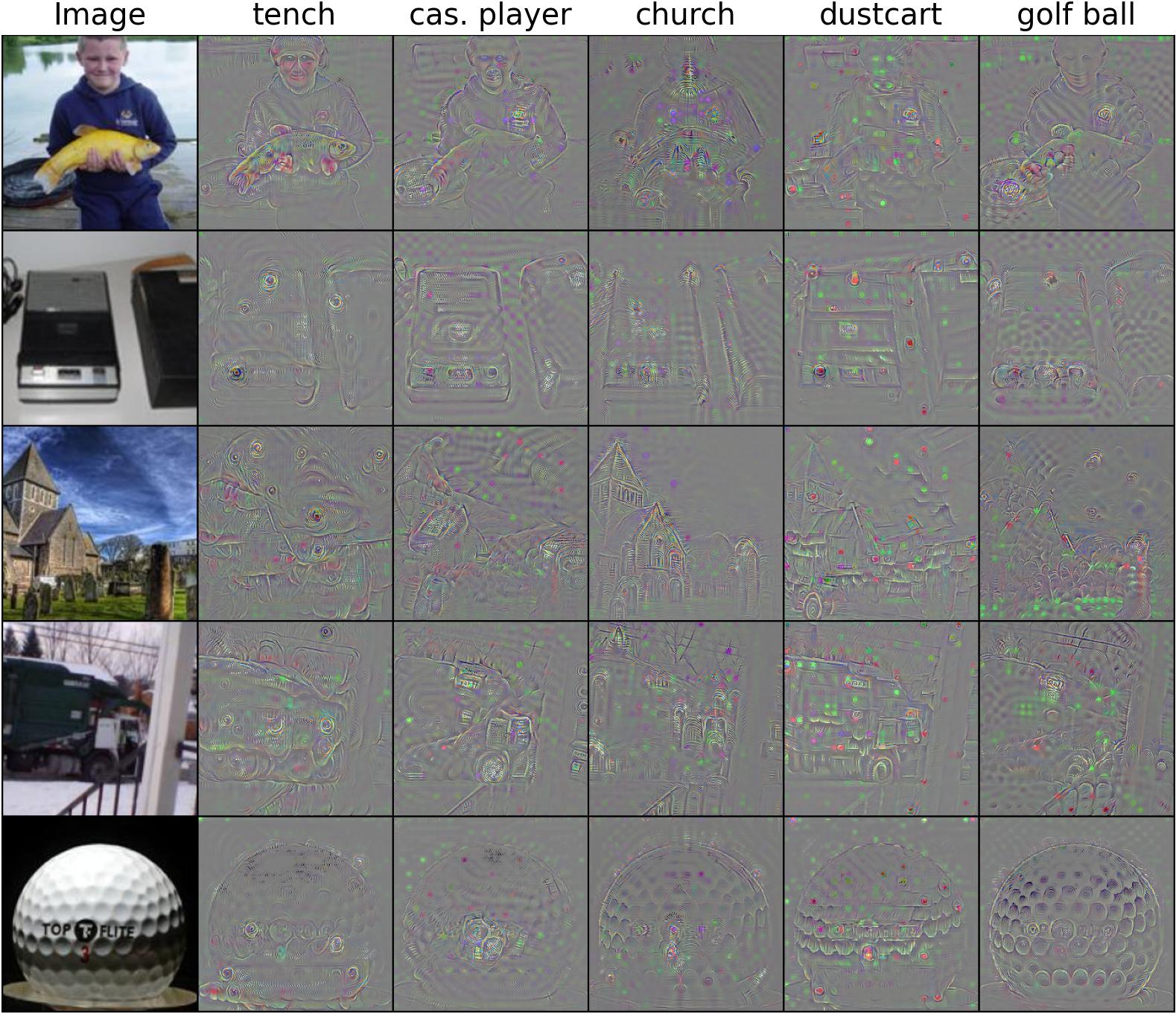}
\hfill
\includegraphics[width=0.49\linewidth]{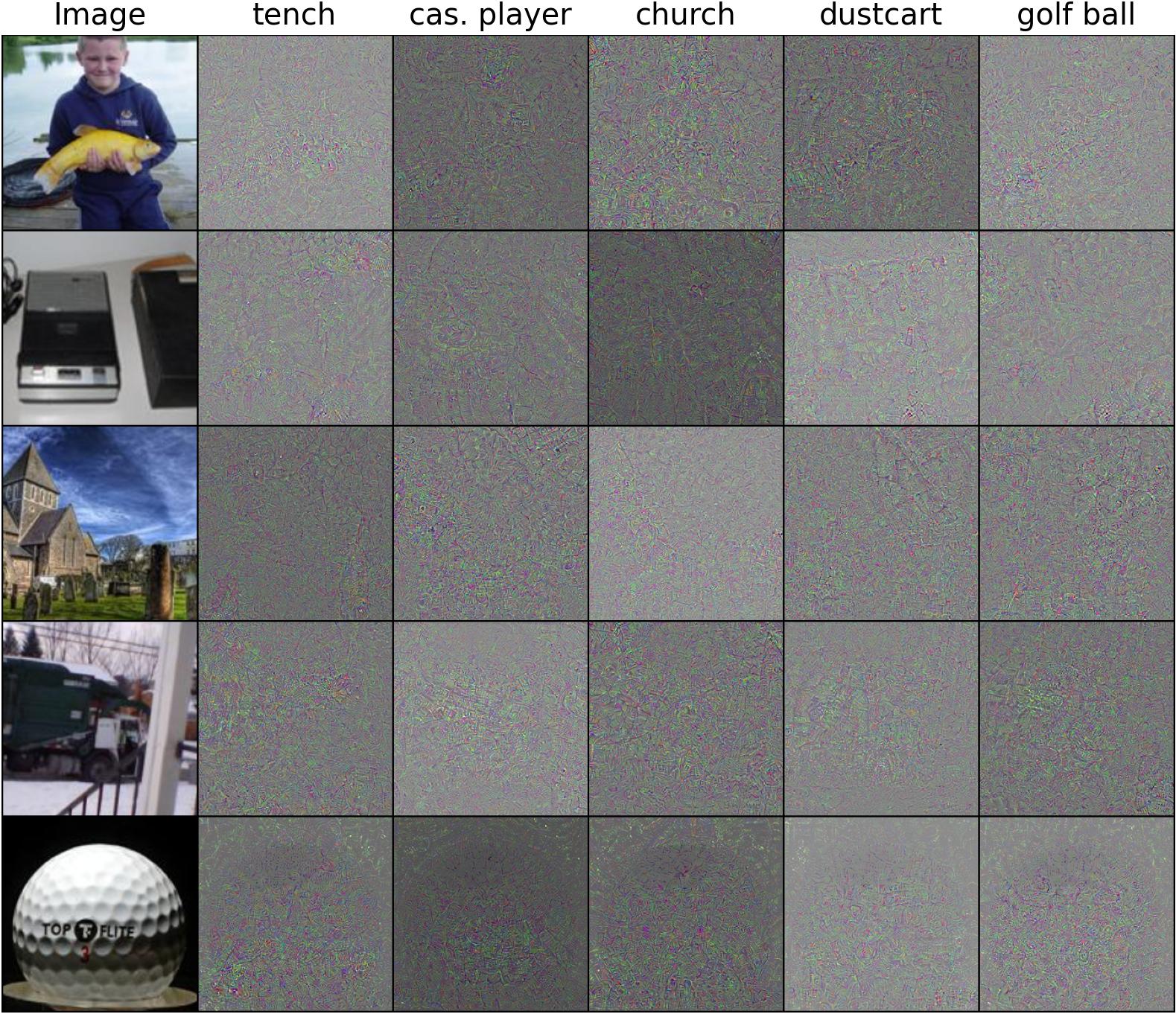}

\caption{Class-specific accentuations for ResNet50 using local ascent with $K=10$ steps.
\textbf{Left}: PullbackAscent produces sharp, perceptually aligned, and target-specific perturbations \emph{without any post-processing}.
\textbf{Right}: GradientAscent produces predominantly noisy, adversarial-looking perturbations under the same settings.
Notably, PullbackAscent can reveal features characteristic of the target class even when the corresponding object is absent from the original image.
Comparisons for VGG and PVT are provided in the Appendix (Figures~\ref{fig:counterfactuals_vgg} and \ref{fig:counterfactuals_pvt}, respectively).}
\label{fig:counterfactuals_resnet}
\end{figure}

The same replacement principle naturally extends beyond single-step
attribution. In particular,
\emph{PullbackAscent} replaces the gradient direction in local activation
ascent. 
\textbf{Strikingly, in contrast to the classical activation maximization along gradients}~\citep{olah2017feature,cammarata2020thread:}, \textbf{\emph{PullbackAscent} produces coherent, target-specific local
perturbations with just a few steps and no post-processing} (cf. Figure~\ref{fig:counterfactuals_resnet}).

We use \emph{Semantic Pullbacks} (SP) to denote explanation procedures built
by replacing gradients with (soft) pullbacks. In this work we study the following pairs: Gradient/SoftPullback,
SmoothGrad/SmoothPullback, FusionGrad/FusionPullback, and
GradientAscent/PullbackAscent.
We then evaluate pullback substitution on 1000 ImageNet validation images using ResNet50~\citep{he2015deepresiduallearningimage},
VGG~\citep{simonyan2015deepconvolutionalnetworkslargescale}, and
PVT~\citep{wang2021pyramidvisiontransformerversatile}, computing
six complementary Quantus~\citep{hedstrom2023quantus} metrics that cover
faithfulness, localisation, stability, and target specificity. This evaluation is, to our knowledge, unusually broad for post-hoc attribution work, combining
multiple architectures, methods, and complementary metrics at ImageNet scale.

Across all architectures and procedures, replacing gradients with pullbacks consistently improves key faithfulness, stability, and localisation measures, while remaining efficient computationally and straightforward conceptually.
In particular, we find that PullbackAscent is competitive with more elaborate attribution methods and produces coherent, perceptually aligned and target-specific local perturbations.

Our main contributions are:

\begin{enumerate}
    \item We identify the pullback of a network's input-conditioned effective
    operator as a principled local explanatory object and a direct
    generalization of linear weight-vector explanations.
    \item We provide efficient layer-wise pullback rules for standard
    pretrained convolutional and transformer models, including backward-only
    softening for hard-gated operators.
    \item We provide quantitative and qualitative evidence that these simple
    gradient-pullback substitutions substantially improve standard explanation procedures, making them competitive with more elaborate methods;
    in particular, PullbackAscent produces sharp, perceptually aligned, and
    target-specific local perturbations in only a few steps, with no
    post-processing.
\end{enumerate}

Overall, our results suggest that \textbf{adjoint transport should be treated
as a first-class primitive alongside gradients and might be the right explanatory direction for neural networks}.
As a plug-and-play replacement for gradients, our approach can simplify attribution pipelines or enhance them without architectural changes or overly complex machinery.

%

\section{Preliminaries}\label{sec:preliminaries}

Let $f:\R^d\to\R^C$ be a pretrained network and let $s_c(\vx)$ denote the logit of class $c$.
More generally, we consider a scalar score specified by a selector $\vu\in\R^C$ (e.g., $\vu=\ve_c$, where $\ve_c$ is the standard basis one-hot vector):
\begin{equation}\label{eq:score_def}
s_\vu(\vx) := \langle \vu, f(\vx)\rangle
\end{equation}
We write $f = h_L\circ\cdots\circ h_1$ with intermediate activations $\vx_0:=\vx$ and $\vx_\ell:=h_\ell(\vx_{\ell-1})$.

\subsection{Pullbacks}

Consider a (dynamic) linear function $\mW:\R^d\rightarrow\R^C$. Then it has a natural \emph{adjoint} function $\mW^\top:\R^C\rightarrow\R^d$ as:
\begin{equation}\label{eq:adjoint_transpose}
\langle \vu, \mW \vx\rangle = \vu^\top \mW \vx = (\vu^\top \mW \vx)^\top = \vx^\top \mW^\top \vu = \langle \mW^\top \vu, \vx\rangle 
\end{equation}
where the second equality is because both sides are scalars.
We call the value of the adjoint $\mW^\top \vu$ the \emph{pullback} of $\vu$. Since $\mW$ is dynamic linear, then the pullback is actually a vector field $\nu_\vu(\vx) = \mW(\vx)^\top \vu$ dependent on $\vx\in\R^d$ that represents the action of $\vu$ on $\mW \vx$.

\subsection{Dynamic linear operators and pullbacks}
For many modern layers, the forward computation at a fixed input can be represented as a dynamic affine\footnote{For formal correctness in the presence of biases, an augmented-matrix formulation is given in Appendix~\ref{app:bias_terms}.} map $\vx_\ell = \mW_\ell(\vx_{\ell-1})\,\vx_{\ell-1}$

where $\mW_\ell(\cdot)$ depends on the current forward state (e.g., gating patterns, pooling switches, attention maps,
normalization statistics).
The corresponding network-level dynamic affine operator can be defined as
\begin{equation}\label{eq:dyn_linear_net}
\mW(\vx) := \mW_L(\vx_{L-1})\circ\cdots\circ \mW_1(\vx_0)
\end{equation}
so that $f(\vx) = \mW(\vx)\,\vx$ and, consequently,
\begin{equation}\label{eq:pullback_representation}
s_\vu(\vx) = \langle \vu, f(\vx)\rangle = \langle \vu, \mW(\vx)\,\vx\rangle=\langle \nu_\vu(\vx), \vx\rangle
\end{equation}
Thus, the vector field $\nu_\vu$ represents $s_\vu$ via the dot product and can be considered the neuron’s explanation in much
the same way as the weight vector of a linear filter ``explains'' the behavior of that filter.

\subsection{Pullback vs.\ gradient}
A common attribution baseline is the gradient $\nabla_{\vx} s_\vu(\vx)$, which differentiates through \emph{all} input-dependence.
Pullback is different: it is the adjoint of the \emph{dynamic linear component} and does \emph{not} differentiate through how $\mW_\ell(\cdot)$ depends on the forward state.
Consequently, pullbacks and gradients coincide for some layer classes (e.g., linear/conv and piecewise-linear routing regimes like ReLU or MaxPool), but \emph{not in general} (e.g., SiLU/GELU, LayerNorm, Attention), where gradients include additional terms from
differentiating through gates/statistics/softmax.

\begin{figure}[tb]
  \begin{center}
    \centerline{\includegraphics[width=\columnwidth]{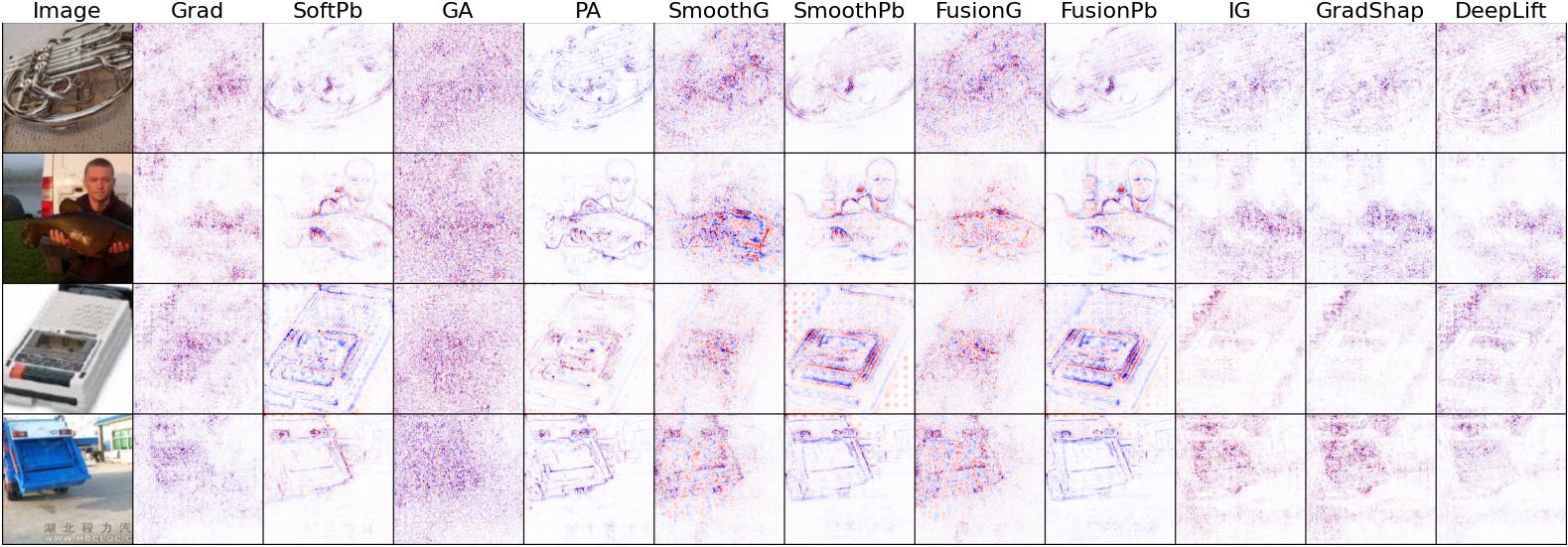}}
    \caption{Pullback-based explainer heatmaps are qualitatively superior to their gradient-based counterparts (comparison for ResNet50). Comparisons for VGG and PVT are provided in the Appendix (Figures~\ref{fig:comparison_cnn}, \ref{fig:comparison_pvt_v2_b1}).}
    \label{fig:comparison_resnet50_4}
  \end{center}
\end{figure}
\section{Method}\label{sec:method}

Given a pretrained network and target selector $\vu$ (typically $\vu=\ve_c$), we seek to recover coherent local directions associated with the target score. We argue that features can be only partially activated and therefore a single backward pass may recover an incomplete input-space pattern.
We address this through two complementary mechanisms: \emph{SoftPullback} (SfP), which softens backward gating of the hard-gated layers (ReLU, MaxPool), and \emph{PullbackAscent} (PA), which iteratively follows the pullback vector field to reinforce locally preferred feature directions.

\subsection{Soft adjoints of layers}\label{sec:soft_adjoint}

Section~\ref{sec:preliminaries} defines the standard pullback as
$\nu_\vu(\vx)=\mW(\vx)^\top \vu$, where $\mW(\vx)$ is the network's dynamic linear operator.
Despite being algebraically faithful
(cf. Equation~\ref{eq:pullback_representation}), this representation can be perceptually noisy when gating or routing inconsistently selects components of a coherent feature.
This issue is particularly pronounced in hard-gated layers:
ReLU completely removes locally negative components, while MaxPool routes the signal exclusively through the winning location.
Thus, small changes in the input can abruptly remove or restore entire components of the explanation.

We soften these hard backward decisions while leaving the forward computation unchanged.
The guiding interpretation is a \emph{local expected pullback}, obtained by averaging the pullback over a neighborhood of the input under the data distribution.

\paragraph{Motivation}
For ReLU after BatchNorm, if preactivations are locally distributed as $Z\sim\mathcal{N}(Z(\vx),\sigma^2)$ around the input $\vx\in\sR^d$, then $\mathbb{E}[\1{Z>0}]=\Phi(Z(\vx)/\sigma)$. Thus, the expected gate is \textit{soft}, which suggests replacing hard backward gating with a softened gate (e.g., Normal CDF with temperature). If we additionally assume that preactivations are locally mutually independent, then the pullback through the softer gating approximates the expected pullback in the input neighborhood, without altering the forward pass.
We offer additional mechanistic motivation in Appendix~\ref{app:appendix_motivation}.

For each layer, we define a \emph{soft adjoint}
$\tilde{\mW}_\ell^\top(\cdot;\tau)$ that preserves the forward computation while allowing backward gating or routing to be softened.
For example, for SiLU, $\phi(z)=z\,\sigma(z)$, the standard adjoint of the dynamic linear component multiplies by $\sigma(z)$ element-wise, whereas the soft adjoint uses $\sigma(z/\tau)$.
Unlike the gradient, neither operator differentiates through the gate.

In architectures with stronger intra-layer dependencies, such as those involving Self-Attention and LayerNorm, local independence assumptions become less plausible, weakening the expected-pullback justification for layer-wise softening.
We therefore concentrate explicit softening on hard-gated layers and set $\tau=1$ for layers with already-soft gating or routing, retaining their original gates.
Appendix~\ref{app:layer_pullbacks} provides explicit layer-wise definitions.

\subsection{Soft Pullback}\label{sec:soft_pullback}

Let
$\tilde{\mW}(\vx;\tau):=
\tilde{\mW}_L(\vx_{L-1};\tau)\cdots\tilde{\mW}_1(\vx_0;\tau)$
denote the composition of the layer operators corresponding to the soft adjoints, evaluated on the cached forward state.
Here, $\tau$ denotes the softening temperature, with temperature fixed to $1$ in already-soft gates.
We define \emph{Soft Pullback} (SfP) as
\begin{equation}\label{eq:soft_pullback}
\tilde{\nu}_\vu(\vx):=\tilde{\mW}(\vx;\tau)^\top \vu.
\end{equation}
Operationally, $\tilde{\nu}_\vu(\vx)$ is computed by propagating $\vu$ backward using the layer-wise soft adjoints, implemented through custom backward rules
(cf. Appendix~\ref{app:appendix_pytorch}).

\subsection{Pullback Ascent}\label{sec:pullback_ascent}

Softening backward gates can recover suppressed components, but a single pullback may still capture only a weak or inconsistent local expression of a coherent feature.
We therefore use a small number of input-space steps to reinforce the directions selected by the target-conditioned, top-down information flow.

The motivation follows from Equation~\ref{eq:pullback_representation}:
$s_\vu(\vx)=\langle\nu_\vu(\vx),\vx\rangle$.
At each input, the pullback acts as the weight vector of the linear representation selected by the network's current state.
If this operator is held fixed, moving in the direction $\nu_\vu(\vx)$ increases its represented score, since
\begin{equation}
\langle\nu_\vu(\vx),\vx+\alpha\nu_\vu(\vx)\rangle
=
s_\vu(\vx)+\alpha\|\nu_\vu(\vx)\|_2^2
\qquad (\alpha>0).
\end{equation}
This suggests iteratively following the pullback direction and recomputing it after each step, allowing the explanation to adapt as the feature becomes more strongly pronounced.

We refine explanations through
\begin{equation}\label{eq:pullback_ascent}
\tilde{\nu}_\vu^*(\vx):=\vx^{(K)}-\vx,
\qquad
\vx^{(t+1)}
:=
\vx^{(t)}
+
\alpha\cdot
\mathrm{Norm}\!\left(\tilde{\nu}_\vu(\vx^{(t)})\right),
\end{equation}
where $\vx^{(0)}=\vx$, $\mathrm{Norm}(\cdot)$ is an optional $\ell_2$ normalization, $\alpha$ is the step size (typically $\alpha=20$), and $K$ is small, e.g., $K\in\{3,5,10\}$, typically $K=5$.
We refer to the resulting displacement $\tilde{\nu}_\vu^*(\vx)$ as \emph{PullbackAscent} (PA). We also consider \emph{GradientAscent} (GA) as a baseline, defined analogously, but replacing the soft pullback direction $\tilde{\nu}_\vu(\vx^{(t)})$ with the standard input gradient $\nabla_{\vx} s_\vu(\vx^{(t)})$.


\subsection{Sampling-based explanations}\label{sec:sampling_explainers}

To complement local saliency and ascent-based explanations, we also replace gradients with pullbacks in sampling-based attribution methods.
Specifically, we consider SmoothGrad and FusionGrad, which aggregate explanations over perturbed versions of the input and the network parameters.
For each sampled input, we replace the standard input gradient with the corresponding pullback, while leaving the sampling distribution and aggregation procedure unchanged.

This yields pullback-based counterparts of SmoothGrad and FusionGrad and allows us to evaluate the effect of replacing gradients with pullbacks within otherwise matched explanation procedures.

\paragraph{Semantic Pullback}
Together with direct saliency and iterative ascent, these experiments cover three natural ways of constructing local explanations: single-point attribution, iterative refinement, and sampling-based aggregation. We use \emph{Semantic Pullback} (SP) as an umbrella term for these refined methods. Notice that \textbf{we don't apply any post-processing or regularization} to the Semantic Pullbacks, in particular no smoothing nor restricting to positive components.

\subsection{Feature accentuation and counterfactual generation}\label{sec:accentuations}
\begin{figure}[tb]
  \begin{center}
    \centerline{\includegraphics[width=\linewidth]{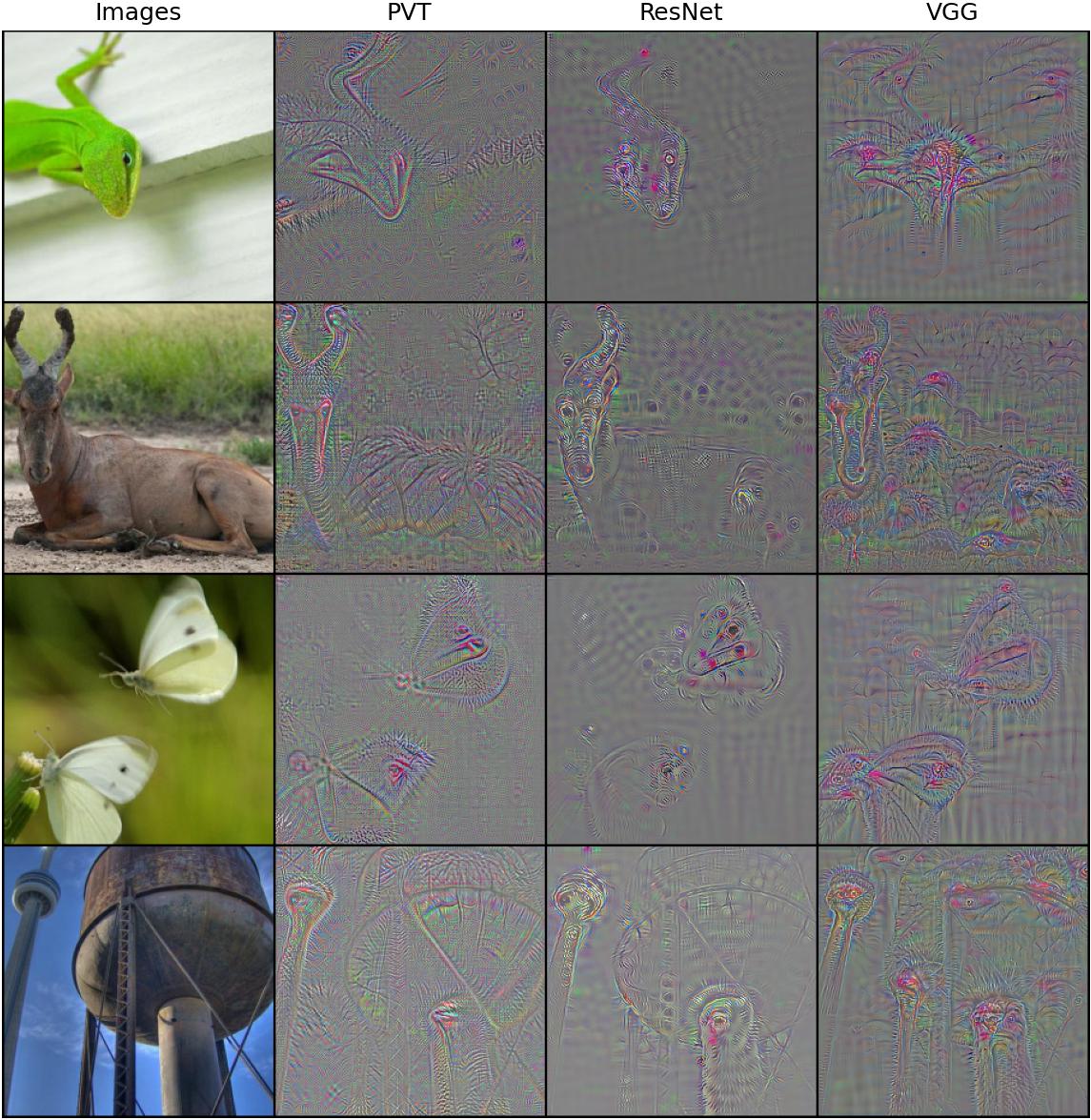}}
    \caption{Accentuations for the class ``ostrich`` using PullbackAscent with K=10 on random ImageNet samples. The ostrich-like features (neck, eyes, beak, fur) are clearly visible despite there being no ostriches in the original images. Further examples are shown in Figures \ref{fig:accentuations_true_label}, \ref{fig:accentuations_mosque}, \ref{fig:accentuations_mud_turtle}, and \ref{fig:accentuations_sidewinder} in the Appendix.}
    \label{fig:targeted_PA}
  \end{center}
\end{figure}

Beyond attribution, any input-space explanation can be interpreted as a local perturbation that accentuates the feature encoded by the target neuron $\vu$.
For a fixed explainer $e$, given the explainer output $e_\vu(\vx)$ for a target $\vu$, we define
\begin{equation}\label{eq:accentuations}
\delta_\vu(\vx) := \vx_\vu - \vx,
\qquad \text{with} \quad
\vx_\vu := \Pi_{\sX}\!\left(\vx + e_\vu(\vx)\right)
\end{equation}
where $\sX$ is the admissible input space and $\Pi_{\sX}$ is the corresponding projection.
For PullbackAscent, we set $e_\vu(\vx)=\tilde{\nu}^{*}_\vu(\vx)$. If $\vu$ represents an arbitrary chosen output class, this lets us view $\delta_\vu(\vx)$ as a \textbf{counterfactual perturbation}. We use this formulation for qualitative comparison between PullbackAscent and GradientAscent (cf. Figures~\ref{fig:counterfactuals_resnet} and \ref{fig:targeted_PA}).

\section{Experiments}\label{sec:experiments}

Our experiments are designed to answer the following questions:

\paragraph{RQ1:~Generality of gradient-to-pullback substitution}
Does replacing gradients with pullbacks improve otherwise matched explanation procedures across different architectures and attribution settings, including direct saliency, iterative ascent, and sampling-based aggregation?

\paragraph{RQ2:~Faithfulness, stability, and localization}
To what extent do pullback-based explanations improve faithfulness, stability, and spatial localization, and how do these benefits vary across explanation procedures and evaluation metrics?

\paragraph{RQ3:~Perceptual alignment of target-specific perturbations}
Does PullbackAscent produce sharper, more perceptually aligned, and more target-specific local perturbations than GradientAscent, making it suitable for feature accentuation and counterfactual generation?

\paragraph{RQ4:~Competitiveness, generality, and efficiency}
Can these simple pullback-based procedures remain competitive with more elaborate established attribution methods while being computationally efficient, conceptually straightforward, and directly applicable to standard pretrained models?

\subsection{Models, datasets, methods and metrics}\label{sec:exp_setup}
\paragraph{Models}
We evaluate three widely used ImageNet-pretrained architectures: ResNet-50~\citep{he2015deepresiduallearningimage}, VGG11~\citep{simonyan2015deepconvolutionalnetworkslargescale}, and Pyramid Vision Transformer (PVT)~\citep{wang2021pyramidvisiontransformerversatile}.
This selection covers both piecewise-linear routing regimes (CNNs) and attention/normalization-driven routing (PVT), which is
a key setting where pullbacks should not be conflated with gradients.
We choose PVT over ViT because its hierarchical, dense spatial representation provides substantially finer input-space resolution for attribution.
In contrast, standard ViTs operate on relatively coarse image patches, making fine-grained attribution difficult without additional architectural modifications or post-processing: for example, B-cos~\citep{Boehle2023BCosAF} introduces a convolutional stem, while DAVE~\citep{wrobel2026davedistributionawareattributionvit} relies on substantial post-processing to obtain spatially refined explanations.
PVT therefore provides a cleaner setting/controlled environment for evaluating the intrinsic spatial structure of pullback-based explanations.
\paragraph{Datasets}
For quantitative evaluation, we use a random selection of 1000 validation images from ImageNet~\citep{imagenet-object-localization-challenge}, which provides a standard large-scale benchmark while keeping the computational cost of multi-metric attribution evaluation tractable.
Unless stated otherwise, all reported quantitative results are averaged over these samples.
For qualitative visualizations, we use Imagenette~\citep{imagenette}, a 10-class subset of ImageNet, which facilitates compact class-conditional comparisons while retaining natural-image complexity.

\paragraph{Methods}
We evaluate pullbacks both through gradient-to-pullback substitutions and against established attribution baselines.
For the experimental comparisons, we consider \emph{Gradient} vs.\ \emph{SoftPullback}, \emph{GradientAscent} vs.\ \emph{PullbackAscent}, and the sampling-based \emph{SmoothGrad}~\citep{smilkov2017smoothgradremovingnoiseadding} and \emph{FusionGrad}~\citep{bykov2021noisegrad} with either gradients or SoftPullbacks as the underlying explainer.
We additionally compare against representative Captum~\citep{kokhlikyan2020captumunifiedgenericmodel} methods: \emph{GradientShap}~\citep{lundberg2017unifiedapproachinterpretingmodel}, \emph{Integrated Gradients}~\citep{sundararajan2017axiomaticattributiondeepnetworks}, and \emph{DeepLift}~\citep{shrikumar2019learningimportantfeaturespropagating}.
These baselines provide a broader reference for whether simple pullback-based procedures remain competitive with more elaborate attribution methods, while GradientAscent also serves as the baseline for feature accentuation.

\paragraph{Hyperparameters}
Unless stated otherwise, all quantitative experiments use the following fixed hyperparameters.
For Semantic Pullbacks, we set the soft-adjoint temperatures to $\tau_{\mathrm{ReLU}}=0.6$ and $\tau_{\mathrm{MaxPool}}=0.3$.
For attention layers, we set $\tau_{\mathrm{Attn}}=1.0$, i.e., we do not apply backward attention softening in the PVT experiments.
For both PullbackAscent and GradientAscent, we enable $\ell_2$ step normalization and use $\alpha=20$ with $K=5$ ascent steps.
We report hyperparameter ablations in Appendix~\ref{app:hyperparameter_sensitivity}.

%
\paragraph{Metrics}
We use the quantitative evaluation protocols available in Quantus~\citep{hedstrom2023quantus}, a state-of-the-art library for benchmarking model explanations, including deletion/insertion-style faithfulness proxies such as Infidelity~\citep{yeh2019infidelitysensitivityexplanations}, Faithfulness Correlation~\citep{ijcai2020p0417}, Monotonicity Correlation~\citep{nguyen2020quantitativeaspectsmodelinterpretability}, as well as sensitivity and target-specificity measures, i.e. Max Sensitivity~\citep{yeh2019infidelitysensitivityexplanations}, Random Logit~\citep{sixt2024explanationsliemodifiedbp}.
We additionally evaluate localization using Focus~\citep{ariasduart2022focusratingxaimethods}, which measures whether attribution mass is concentrated on image regions corresponding to the target class.
Faithfulness metrics quantify to what extent explanations follow the predictive behaviour of the model in various ways.
We place particular emphasis on Infidelity, which is especially well suited to image data, because it explicitly accounts for the directional structure of the visual features, such as correlated pixels forming edges, textures, and object parts.
We also emphasize Focus as a measure of spatial localization.
For the remaining faithfulness metrics, we modify the Quantus default parameters where appropriate to better match the characteristics of image data, including patch-based perturbations instead of randomly selected individual pixels for Faithfulness Correlation.
Formal definitions of all metrics and deviations from Quantus defaults are provided in Appendix~\ref{app:metrics}.
We use Random Logit as a sanity check for target specificity.

\begin{table}[tb]
\caption{ResNet50 results on 1000 ImageNet validation samples;
Focus is evaluated separately on 1000 ImageNet-derived mosaics.
Entries report means with standard deviations as subscripts.
The first eight methods form four gradient-pullback pairs;
pullback-based method names are bold.
The remaining methods provide context.
The best mean in each of the first five metric columns is bold.
Red entries indicate that the 95\% bootstrap confidence interval
for the pullback-minus-gradient mean difference lies entirely on the
deterioration side of zero; blue entries indicate that the interval
includes zero, providing inconclusive evidence of improvement or
deterioration (see Table~\ref{table:paired_explainer_differences}).
Random Logit reports the absolute value of mean with the
standard deviation of the original signed scores and is excluded
from the ranking and colour annotations.
For PVT and VGG see Tables~\ref{table:pvt_results} and~\ref{table:vgg_results}, respectively.
}
\label{table:resnet50_results}
\begin{center}
\begin{tabular}{lrrrrrr}
\toprule
Method
& Infidelity $\downarrow$
& Focus $\uparrow$
& Faith.Corr $\uparrow$
& Mono.Corr $\uparrow$
& Max.Sens $\downarrow$
& Rand.Logit $\downarrow$ \\
\midrule
Grad
& $83.294_{\pm 68.15}$ & $0.634_{\pm 0.14}$ & $0.241_{\pm 0.36}$ & $0.275_{\pm 0.39}$ & $0.952_{\pm 0.17}$ & $0.032_{\pm 0.23}$ \\
\textbf{SoftPb}
& $\phantom{0}5.989_{\pm 05.67}$ & $\best{0.746}_{\pm 0.13}$ & $\best{0.389}_{\pm 0.37}$ & $0.420_{\pm 0.39}$ & $\best{0.119}_{\pm 0.04}$ & $0.062_{\pm 0.42}$ \\
\pairsep
GA
& $29.480_{\pm 35.36}$ & $0.584_{\pm 0.10}$ & $0.278_{\pm 0.37}$ & $0.231_{\pm 0.32}$ & $1.267_{\pm 0.04}$ & $0.002_{\pm 0.05}$ \\
\textbf{PA}
& $\phantom{0}\best{5.384}_{\pm 04.83}$ & $0.726_{\pm 0.13}$ & $0.382_{\pm 0.37}$ & $\best{0.421}_{\pm 0.40}$ & $0.244_{\pm 0.09}$ & $0.212_{\pm 0.19}$ \\
\pairsep
SmoothG
& $66.883_{\pm 58.16}$ & $0.640_{\pm 0.12}$ & $0.331_{\pm 0.40}$ & $0.364_{\pm 0.36}$ & $0.639_{\pm 0.08}$ & $0.024_{\pm 0.20}$ \\
\textbf{SmoothPb}
& $\phantom{0}8.122_{\pm 10.13}$ & $0.713_{\pm 0.16}$ & $\inconclusive{0.326_{\pm 0.39}}$ & $\inconclusive{0.352_{\pm 0.40}}$ & $0.220_{\pm 0.05}$ & $0.065_{\pm 0.40}$ \\
\pairsep
FusionG
& $58.514_{\pm 57.80}$ & $0.634_{\pm 0.12}$ & $0.344_{\pm 0.39}$ & $0.357_{\pm 0.35}$ & $0.855_{\pm 0.09}$ & $0.014_{\pm 0.17}$ \\
\textbf{FusionPb}
& $10.131_{\pm 14.55}$ & $0.691_{\pm 0.16}$ & $\deterioration{0.306_{\pm 0.40}}$ & $\inconclusive{0.350_{\pm 0.40}}$ & $0.362_{\pm 0.08}$ & $0.051_{\pm 0.39}$ \\
\midrule
IG
& $66.830_{\pm 61.21}$ & $0.626_{\pm 0.17}$ & $0.244_{\pm 0.38}$ & $0.350_{\pm 0.41}$ & $0.788_{\pm 0.21}$ & $0.029_{\pm 0.23}$ \\
GradShap
& $69.072_{\pm 62.93}$ & $0.619_{\pm 0.17}$ & $0.233_{\pm 0.38}$ & $0.344_{\pm 0.41}$ & $1.400_{\pm 0.41}$ & $0.023_{\pm 0.24}$ \\
DeepLift
& $73.188_{\pm 62.60}$ & $0.622_{\pm 0.18}$ & $0.201_{\pm 0.36}$ & $0.324_{\pm 0.42}$ & $0.982_{\pm 0.18}$ & $0.033_{\pm 0.24}$ \\
\bottomrule
\end{tabular}
\end{center}
\end{table}

\subsection{Results discussion}

Quantitative evaluation of explanations remains an open problem, particularly for visual features whose spatial structure and inductive biases may not be adequately captured by correlation-based metrics.
We therefore interpret the quantitative results alongside qualitative comparisons, rather than treating any individual metric as a definitive measure of explanation quality.

\paragraph{Paired comparisons and uncertainty}
The large per-image standard deviations motivate assessing uncertainty in the differences directly. 
For each gradient-pullback pair, we compute per-example score differences (pullback minus gradient) and estimate 95\% percentile bootstrap confidence intervals for their mean using 10,000 resamples.

\paragraph{RQ1:~Generality of gradient-to-pullback substitution}
Across three architectures and four gradient-pullback pairs, replacing gradients with pullbacks yields improvements supported by 95\% bootstrap confidence intervals in 46 of 60 comparisons (Table~\ref{table:paired_explainer_differences}).
Seven comparisons indicate deterioration, while seven remain inconclusive.
The improvements span direct saliency, iterative ascent, and sampling-based aggregation, suggesting that the benefits arise from the underlying backward operator rather than a particular explanation procedure.

Importantly, these improvements occur in two distinct architectural regimes.
For the ReLU-based ResNet50 and VGG, standard pullbacks coincide with gradients, making backward-only softening essential to obtaining different explanations.
Across these CNNs, 33 of 40 comparisons support improvement, suggesting that softening hard backward decisions can recover faithful feature components otherwise suppressed by gating.

For PVT, pullbacks already differ from gradients in attention- and normalization-based operators because they do not differentiate through the input dependence of the effective operator.
Here, 13 of 20 comparisons support improvement, while five indicate deterioration and two remain inconclusive.
This provides evidence that adjoint transport is useful beyond piecewise-linear networks, although its benefits are less uniform in attention-based architectures.

\paragraph{RQ2:~Faithfulness, stability, and localization}
The strongest quantitative evidence concerns Infidelity and Focus, with all 12 matched comparisons supporting improvement for each metric.
The consistently favorable Infidelity results suggest that pullback-based explanations better capture input-space directions associated with the model's effective action (cf.~Equation~\ref{eq:pullback_representation}), while Focus improvements indicate better localization of target-relevant features.
Max Sensitivity also favors pullbacks in 11 of 12 comparisons, supporting improved local stability.
In contrast, correlation-based faithfulness metrics yield more mixed evidence, particularly Monotonicity Correlation.

PullbackAscent stands out across these criteria: its paired confidence intervals support improvement over GradientAscent in 14 of 15 comparisons, with PVT Faithfulness Correlation remaining inconclusive.
It achieves the lowest Infidelity among all evaluated methods, reducing it by factors of 5.5, 2.8 and 3.3  relative to GradientAscent on ResNet50, PVT and VGG, respectively.
This suggests that iterative refinement can substantially strengthen the explanatory value of pullback directions.

The relative strengths of individual procedures nevertheless differ across architectures.
For the ReLU-based CNNs, single-step SoftPullback already provides substantial improvements, suggesting that softening backward gates alone recovers much of the useful local feature structure.
For PVT, the benefits are more metric-dependent, with iterative ascent and sampling-based aggregation offering complementary advantages.
Monotonicity Correlation remains an exception, particularly on PVT and VGG.

These quantitative findings are complemented by the qualitative attribution comparisons in Figures~\ref{fig:comparison_resnet50_4}, \ref{fig:comparison_cnn} and \ref{fig:comparison_pvt_v2_b1}, where pullback-based methods generally produce more coherent and spatially localized explanations than their gradient-based counterparts.
Given the limitations of numerical attribution metrics, this qualitative consistency constitutes an important part of the empirical evidence.

\paragraph{RQ3:~Perceptual alignment of target-specific perturbations}
PullbackAscent provides a particularly striking qualitative demonstration of the benefits of replacing gradients with pullbacks.
While GradientAscent usually produces noisy or adversarial-looking perturbations, PullbackAscent generates sharp, perceptually aligned, and target-specific structures in only a few steps, without post-processing or perceptual priors.

These perturbations accentuate existing visual structures in directions associated with the target class, making PullbackAscent well suited to feature accentuation and counterfactual generation (cf.~Figures~\ref{fig:counterfactuals_resnet}, \ref{fig:targeted_PA} and Figures~\ref{fig:counterfactuals_vgg}, \ref{fig:counterfactuals_pvt},~\ref{fig:accentuations_true_label}, \ref{fig:accentuations_mosque}, \ref{fig:accentuations_mud_turtle}, \ref{fig:accentuations_sidewinder} in the Appendix).
Importantly, these effects are observed in both ReLU-based CNNs and PVT, despite the different mechanisms underlying their pullback operators.
The strong Infidelity and Focus results provide complementary quantitative evidence that the resulting perturbations reflect the model's predictive behavior and target-relevant spatial structure.

An interesting exception concerns Random Logit, where PullbackAscent often performs less favorably.
Although commonly interpreted as a sanity check for class sensitivity, this metric requires caution for iteratively generated, target-specific perturbations.
Our visualizations suggest that PullbackAscent can reuse features shared across classes and progressively transform existing structures toward the selected target, even when that class is absent from the original image.
A higher Random Logit score may therefore partly reflect shared visual structure, instead of a lack of target specificity.

More generally, these observations suggest that PullbackAscent does more than merely amplify existing saliency: by repeatedly following target-conditioned pullback directions, it can progressively reshape the local expression of features toward a selected class.
This behavior is especially relevant to counterfactual generation, where the goal is not only to localize existing evidence but also to identify minimal yet meaningful target-directed changes to the input.

\paragraph{RQ4:~Competitiveness, generality, and efficiency}
Beyond the comparisons, pullback-based methods also outperform the more elaborate established attribution baselines in our evaluation.
Across the five main metrics and three architectures, a pullback-based method achieves the best mean in 13 out of 15 cases.
The only exceptions are Monotonicity Correlation on VGG and PVT, where SmoothGrad performs best.
Notably, neither Integrated Gradients, GradientShap, nor DeepLift achieves the best result in any of these comparisons.

These findings suggest that improving the underlying backward operator can be more consequential than introducing increasingly elaborate explanation procedures.
In particular, simple schemes such as direct saliency, iterative ascent, and sampling-based aggregation can achieve strong attribution quality when equipped with pullbacks instead of gradients.
This is encouraging because these procedures are conceptually straightforward, while the underlying pullback operator has a direct interpretation through the adjoint of the network's effective linear representation.
Our results therefore suggest that more complex attribution schemes are not necessarily required to obtain high-quality explanations, at least within the evaluated architectures and metrics.

The results also support architectural generality.
Pullback-based explanations improve standard procedures in both hard-gated convolutional networks and attention-based transformers, despite the different relationships between pullbacks and gradients in these architectures.
All methods operate on standard pretrained models, without retraining or changes to the forward computation.

Computationally, a single SoftPullback requires approximately one backward pass, with modest overhead from the layer-specific backward rules.
PullbackAscent scales approximately linearly with the number of steps $K$, which remains small in practice.
Both avoid the repeated stochastic evaluations required by sampling-based explanations, while retaining competitive quantitative and qualitative performance.
The sampling-based pullback variants preserve the computational structure of their corresponding gradient-based procedures.
Empirical wall-clock measurements in Table~\ref{table:time_complexity} further demonstrate the favorable efficiency of the proposed methods.

\paragraph{Ablations}
Our exploratory comparisons of sigmoid and Normal-CDF surrogate gates suggest qualitatively similar behaviour, while moderate increases in ascent strength or step count tend to make target-specific features more pronounced.
The quantitative sweeps reveal trade-offs on either side of the defaults: additional steps can improve Infidelity on PVT and VGG while reducing Focus or correlation-based faithfulness; smaller step sizes can improve localisation and, on CNNs, stability. The defaults therefore balance perceptual clarity and quantitative performance rather than optimise every metric. Our ablations are described more in-depth in Appendix~\ref{app:hyperparameter_sensitivity}.

Overall, our experiments demonstrate that replacing gradients with pullbacks provides a simple, mathematically motivated, and broadly applicable way to improve standard explanation procedures. Across convolutional and transformer-based models, pullback-based methods deliver consistent gains in Infidelity, localization, and stability, while frequently outperforming more elaborate established baselines. PullbackAscent further demonstrates that following the pullback vector field can generate sharp, perceptually aligned, and target-specific perturbations in only a few steps, without post-processing or strong structural priors.

\section{Related work and positioning}\label{sec:positioning}

\subsection{Gradient-based explanations}\label{sec:gradient_based_explanations}

Gradient-based attribution methods are widely used to explain neural network predictions~\citep{simonyan2014deepinsideconvolutionalnetworks, szegedy2014intriguingpropertiesneuralnetworks}, but often produce noisy or unstable explanations~\citep{mahendran2014understandingdeepimagerepresentations, smilkov2017smoothgradremovingnoiseadding}.
Their reliability has been questioned through sanity checks, studies of causal interpretability, and adversarial manipulation~\citep{ghorbani2018interpretationneuralnetworksfragile, adebayo2020sanitycheckssaliencymaps, zimmermann2021featurevisualizationssupportcausal, geirhos2024donttrusteyesunreliability, Deng2025AttributionEF, heo2019foolingneuralnetworkinterpretations, dombrowski2019explanationsmanipulatedgeometryblame, Baniecki_2024}.
Several established methods, including Integrated Gradients~\citep{sundararajan2017axiomaticattributiondeepnetworks}, GradientShap~\citep{lundberg2017unifiedapproachinterpretingmodel}, and DeepLift~\citep{shrikumar2019learningimportantfeaturespropagating}, construct attributions through path integration, baseline sampling, and reference-based contribution propagation, respectively.

In contrast, our work emphasizes a conceptual distinction between \emph{gradients} and the \emph{adjoint action} that pulls a neuron's linear functional back to input space.
From the dynamic affine viewpoint, the layer adjoint is as natural an operation as the derivative, yet it is rarely exposed explicitly in current libraries.
Notably, this requires specialized backward rules only for a relatively small catalogue of mechanisms that introduce gating, routing, or normalization effects, while coinciding with standard backpropagation for linear components, including convolutions and residual connections.

\subsection{Backward-pass modifications and surrogate gradients}

Alternative backward computations have been explored to improve interpretability without modifying the forward model.
Modified backward slopes can facilitate activation maximization~\citep{Linse_2024}, while surrogate-gradient methods allow backward operators to differ from exact derivatives~\citep{otte2024flexibleefficientsurrogategradient}, sometimes even improving generalisation~\citep{horuz2025resurrectionrelu}.

Our approach is related but mathematically distinct: adjoints are motivated by the effective linear operator, while additional backward-only softening aims to approximate its local expected value.

\subsection{Effective operators and interpretability by design}

B-cos networks encourage alignment between effective weights (which correspond to standard pullbacks) and inputs~\citep{Boehle2023BCosAF}, while B-cosification extends this principle to pretrained models through architectural conversion and fine-tuning~\citep{arya2025bcosificationtransformingdeepneural}.

Our results suggest that meaningful weight-input alignment is already present in standard pretrained networks, without architectural modification or retraining.
A closely related perspective is provided by DAVE~\citep{wrobel2026davedistributionawareattributionvit}, which essentially refines standard pullbacks for vision transformers via averaging over input perturbations and spatial transformations.

\subsection{Gradient smoothing}

SmoothGrad~\citep{smilkov2017smoothgradremovingnoiseadding} and FusionGrad~\citep{bykov2021noisegrad} improve explanation stability through stochastic averaging over inputs or network parameters.
These methods suggest that meaningful explanatory directions may emerge more clearly from local averages than from pointwise gradients.

Semantic Pullbacks pursue a related intuition through closed-form softening of hard-gating decisions, avoiding repeated stochastic evaluations (cf.~Table~\ref{table:time_complexity}).
Smoothing and pullback substitution are also complementary, as demonstrated by SmoothPullback and FusionPullback, while PullbackAscent achieves particularly strong Infidelity results through iterative refinement, providing another effective way to exploit the local directional structure exposed by pullbacks.

\subsection{Feature visualization}

Feature visualization has a long tradition as a practical tool for probing what internal units respond to, including optimization-based methods and circuit-style analysis~\citep{olah2017feature, cammarata2020thread:}.
More recently, Feature Accentuation produces input-conditioned visualizations intended to reveal semantically coherent features in natural images~\citep{hamblin2024featureaccentuationrevealingwhat}, but typically relies on strong regularization and iterative optimization.

PullbackAscent instead produces sharp, target-specific perturbations in only a few steps without heavy post-processing or explicit image priors, suggesting that perceptually meaningful structures can emerge from the choice of backward operator itself.
This positions SP as a suitable lightweight probe of input-space feature structure across convolutional and transformer-based models.

\section{Conclusion, limitations and future work}\label{sec:limitations}

We introduced \emph{Semantic Pullbacks}, a family of explanation procedures based on the adjoint of a network's input-conditioned effective operator, instead of its full input gradient.
Across convolutional and transformer-based architectures, replacing gradients with pullbacks consistently improves key faithfulness, localization, and stability measures, often outperforming more elaborate attribution methods.
Particularly striking is PullbackAscent, which produces sharp, perceptually aligned, and target-specific perturbations in only a few steps, without post-processing or explicit perceptual priors.
Together, these findings suggest that the choice of the underlying backward operator can be more consequential than the complexity of the explanation procedure itself.

Several limitations remain.
Although we evaluate multiple architectures and complementary quantitative metrics, explanation quality lacks a universally accepted evaluation protocol, and the relationship between quantitative faithfulness and perceptual interpretability warrants further investigation.
Our experiments are limited to image classification, while the applicability of pullbacks to language and multimodal models remains unexplored.
Moreover, our implementation uses shared softening temperatures and does not explicitly model bias contributions (cf.~Appendix~\ref{app:bias_robustness}).

Future work includes adaptive layer- or neuron-specific softening, improved pullback rules for attention-based architectures, and extensions to hidden neurons and other modalities.
More broadly, the effective-operator perspective motivates studying feature flow through network paths (cf.~Appendix~\ref{app:path_expansion}), with potential applications to architecture diagnostics, pruning, and counterfactual generation.

\subsubsection*{Author Contributions}

Maciej Satkiewicz led the project, including the conception of the method,
theoretical development and framing, implementation, experimental design and
evaluation, visualization of results, analysis, and manuscript writing. Roberto
Corizzo and Marcin Pietroń contributed through discussions, feedback, manuscript
revisions, suggestions regarding the choice of models, evaluation metrics, and
experimental design, as well as assistance in describing and interpreting the
experimental results.

\subsubsection*{Acknowledgments}

Maciej Satkiewicz gratefully acknowledges Benjamin Crouzier (Tufa Labs) for supporting this research through a research grant awarded to the 314 Foundation.


\clearpage
\medskip

\bibliography{references}

@misc{imagenette,
  author       = {Howard, Jeremy},
  title        = {Imagenette and Imagewoof: Subsets of ImageNet for quick experiments},
  year         = {2019},
  howpublished = {\url{https://github.com/fastai/imagenette}},
  note         = {Accessed: 2025-07-21}
}

@misc{lakshminarayanan2021neuralpathfeaturesneural,
      title={Neural Path Features and Neural Path Kernel : Understanding the role of gates in deep learning}, 
      author={Chandrashekar Lakshminarayanan and Amit Vikram Singh},
      year={2021},
      eprint={2006.10529},
      archivePrefix={arXiv},
      primaryClass={cs.LG},
      url={https://arxiv.org/abs/2006.10529}, 
}

@misc{hamblin2024featureaccentuationrevealingwhat,
      title={Feature Accentuation: Revealing 'What' Features Respond to in Natural Images}, 
      author={Chris Hamblin and Thomas Fel and Srijani Saha and Talia Konkle and George Alvarez},
      year={2024},
      eprint={2402.10039},
      archivePrefix={arXiv},
      primaryClass={cs.CV},
      url={https://arxiv.org/abs/2402.10039}, 
}

@inproceedings{Linse_2024,
   title={Leaky ReLUs That Differ in Forward and Backward Pass Facilitate Activation Maximization in Deep Neural Networks},
   url={http://dx.doi.org/10.1109/IJCNN60899.2024.10650881},
   DOI={10.1109/ijcnn60899.2024.10650881},
   booktitle={2024 International Joint Conference on Neural Networks (IJCNN)},
   publisher={IEEE},
   author={Linse, Christoph and Barth, Erhardt and Martinetz, Thomas},
   year={2024},
   month=jun, pages={1–8} }

@misc{horuz2025resurrectionrelu,
      title={The Resurrection of the ReLU}, 
      author={Coşku Can Horuz and Geoffrey Kasenbacher and Saya Higuchi and Sebastian Kairat and Jendrik Stoltz and Moritz Pesl and Bernhard A. Moser and Christoph Linse and Thomas Martinetz and Sebastian Otte},
      year={2025},
      eprint={2505.22074},
      archivePrefix={arXiv},
      primaryClass={cs.LG},
      url={https://arxiv.org/abs/2505.22074}, 
}

@misc{he2015deepresiduallearningimage,
      title={Deep Residual Learning for Image Recognition}, 
      author={Kaiming He and Xiangyu Zhang and Shaoqing Ren and Jian Sun},
      year={2015},
      eprint={1512.03385},
      archivePrefix={arXiv},
      primaryClass={cs.CV},
      url={https://arxiv.org/abs/1512.03385}, 
}

@misc{simonyan2015deepconvolutionalnetworkslargescale,
      title={Very Deep Convolutional Networks for Large-Scale Image Recognition}, 
      author={Karen Simonyan and Andrew Zisserman},
      year={2015},
      eprint={1409.1556},
      archivePrefix={arXiv},
      primaryClass={cs.CV},
      url={https://arxiv.org/abs/1409.1556}, 
}

@misc{otte2024flexibleefficientsurrogategradient,
      title={Flexible and Efficient Surrogate Gradient Modeling with Forward Gradient Injection}, 
      author={Sebastian Otte},
      year={2024},
      eprint={2406.00177},
      archivePrefix={arXiv},
      primaryClass={cs.LG},
      url={https://arxiv.org/abs/2406.00177}, 
}

@article{cammarata2020thread:,
  author = {Cammarata, Nick and Carter, Shan and Goh, Gabriel and Olah, Chris and Petrov, Michael and Schubert, Ludwig and Voss, Chelsea and Egan, Ben and Lim, Swee Kiat},
  title = {Thread: Circuits},
  journal = {Distill},
  year = {2020},
  note = {https://distill.pub/2020/circuits},
  doi = {10.23915/distill.00024}
}

@article{Deng2025AttributionEF,
  title={Attribution Explanations for Deep Neural Networks: A Theoretical Perspective},
  author={Huiqi Deng and Hongbin Pei and Quanshi Zhang and Mengnan Du},
  journal={ArXiv},
  year={2025},
  volume={abs/2508.07636},
  url={https://api.semanticscholar.org/CorpusID:280566749}
}

@misc{smilkov2017smoothgradremovingnoiseadding,
      title={SmoothGrad: removing noise by adding noise}, 
      author={Daniel Smilkov and Nikhil Thorat and Been Kim and Fernanda Viégas and Martin Wattenberg},
      year={2017},
      eprint={1706.03825},
      archivePrefix={arXiv},
      primaryClass={cs.LG},
      url={https://arxiv.org/abs/1706.03825}, 
}

@misc{adebayo2020sanitycheckssaliencymaps,
      title={Sanity Checks for Saliency Maps}, 
      author={Julius Adebayo and Justin Gilmer and Michael Muelly and Ian Goodfellow and Moritz Hardt and Been Kim},
      year={2020},
      eprint={1810.03292},
      archivePrefix={arXiv},
      primaryClass={cs.CV},
      url={https://arxiv.org/abs/1810.03292}, 
}

@article{Baniecki_2024,
   title={Adversarial attacks and defenses in explainable artificial intelligence: A survey},
   volume={107},
   ISSN={1566-2535},
   url={http://dx.doi.org/10.1016/j.inffus.2024.102303},
   DOI={10.1016/j.inffus.2024.102303},
   journal={Information Fusion},
   publisher={Elsevier BV},
   author={Baniecki, Hubert and Biecek, Przemyslaw},
   year={2024},
   month=jul, pages={102303} }

@misc{simonyan2014deepinsideconvolutionalnetworks,
      title={Deep Inside Convolutional Networks: Visualising Image Classification Models and Saliency Maps}, 
      author={Karen Simonyan and Andrea Vedaldi and Andrew Zisserman},
      year={2014},
      eprint={1312.6034},
      archivePrefix={arXiv},
      primaryClass={cs.CV},
      url={https://arxiv.org/abs/1312.6034}, 
}

@article{olah2017feature,
  author = {Olah, Chris and Mordvintsev, Alexander and Schubert, Ludwig},
  title = {Feature Visualization},
  journal = {Distill},
  year = {2017},
  note = {https://distill.pub/2017/feature-visualization},
  doi = {10.23915/distill.00007}
}

@misc{ghorbani2018interpretationneuralnetworksfragile,
      title={Interpretation of Neural Networks is Fragile}, 
      author={Amirata Ghorbani and Abubakar Abid and James Zou},
      year={2018},
      eprint={1710.10547},
      archivePrefix={arXiv},
      primaryClass={stat.ML},
      url={https://arxiv.org/abs/1710.10547}, 
}

@misc{lundberg2017unifiedapproachinterpretingmodel,
      title={A Unified Approach to Interpreting Model Predictions}, 
      author={Scott Lundberg and Su-In Lee},
      year={2017},
      eprint={1705.07874},
      archivePrefix={arXiv},
      primaryClass={cs.AI},
      url={https://arxiv.org/abs/1705.07874}, 
}

@misc{zimmermann2021featurevisualizationssupportcausal,
      title={How Well do Feature Visualizations Support Causal Understanding of CNN Activations?}, 
      author={Roland S. Zimmermann and Judy Borowski and Robert Geirhos and Matthias Bethge and Thomas S. A. Wallis and Wieland Brendel},
      year={2021},
      eprint={2106.12447},
      archivePrefix={arXiv},
      primaryClass={cs.CV},
      url={https://arxiv.org/abs/2106.12447}, 
}

@misc{geirhos2024donttrusteyesunreliability,
      title={Don't trust your eyes: on the (un)reliability of feature visualizations}, 
      author={Robert Geirhos and Roland S. Zimmermann and Blair Bilodeau and Wieland Brendel and Been Kim},
      year={2024},
      eprint={2306.04719},
      archivePrefix={arXiv},
      primaryClass={cs.CV},
      url={https://arxiv.org/abs/2306.04719}, 
}

@article{Boehle2023BCosAF,
  title={B-Cos Alignment for Inherently Interpretable CNNs and Vision Transformers},
  author={Moritz D Boehle and Navdeeppal Singh and Mario Fritz and Bernt Schiele},
  journal={IEEE Transactions on Pattern Analysis and Machine Intelligence},
  year={2023},
  volume={46},
  pages={4504-4518},
  url={https://api.semanticscholar.org/CorpusID:259203376}
}

@misc{arya2025bcosificationtransformingdeepneural,
      title={B-cosification: Transforming Deep Neural Networks to be Inherently Interpretable}, 
      author={Shreyash Arya and Sukrut Rao and Moritz Böhle and Bernt Schiele},
      year={2025},
      eprint={2411.00715},
      archivePrefix={arXiv},
      primaryClass={cs.CV},
      url={https://arxiv.org/abs/2411.00715}, 
}

@article{hedstrom2023quantus,
  author  = {Anna Hedstr{\"{o}}m and Leander Weber and Daniel Krakowczyk and Dilyara Bareeva and Franz Motzkus and Wojciech Samek and Sebastian Lapuschkin and Marina Marina M.{-}C. H{\"{o}}hne},
  title   = {Quantus: An Explainable AI Toolkit for Responsible Evaluation of Neural Network Explanations and Beyond},
  journal = {Journal of Machine Learning Research},
  year    = {2023},
  volume  = {24},
  number  = {34},
  pages   = {1--11},
  url     = {http://jmlr.org/papers/v24/22-0142.html}
}

@inproceedings{ijcai2020p0417,
  title     = {Evaluating and Aggregating Feature-based Model Explanations},
  author    = {Bhatt, Umang and Weller, Adrian and Moura, José M. F.},
  booktitle = {Proceedings of the Twenty-Ninth International Joint Conference on
               Artificial Intelligence, {IJCAI-20}},
  publisher = {International Joint Conferences on Artificial Intelligence Organization},
  editor    = {Christian Bessiere},
  pages     = {3016--3022},
  year      = {2020},
  month     = {7},
  note      = {Main track},
  doi       = {10.24963/ijcai.2020/417},
  url       = {https://doi.org/10.24963/ijcai.2020/417},
}

@misc{nguyen2020quantitativeaspectsmodelinterpretability,
      title={On quantitative aspects of model interpretability}, 
      author={{A}n-phi Nguyen and María Rodríguez Martínez},
      year={2020},
      eprint={2007.07584},
      archivePrefix={arXiv},
      primaryClass={cs.LG},
      url={https://arxiv.org/abs/2007.07584}, 
}

@misc{yeh2019infidelitysensitivityexplanations,
      title={On the (In)fidelity and Sensitivity for Explanations}, 
      author={Chih-Kuan Yeh and Cheng-Yu Hsieh and Arun Sai Suggala and David I. Inouye and Pradeep Ravikumar},
      year={2019},
      eprint={1901.09392},
      archivePrefix={arXiv},
      primaryClass={cs.LG},
      url={https://arxiv.org/abs/1901.09392}, 
}

@misc{sixt2024explanationsliemodifiedbp,
      title={When Explanations Lie: Why Many Modified BP Attributions Fail}, 
      author={Leon Sixt and Maximilian Granz and Tim Landgraf},
      year={2024},
      eprint={1912.09818},
      archivePrefix={arXiv},
      primaryClass={cs.LG},
      url={https://arxiv.org/abs/1912.09818}, 
}

@misc{kokhlikyan2020captumunifiedgenericmodel,
      title={Captum: A unified and generic model interpretability library for PyTorch}, 
      author={Narine Kokhlikyan and Vivek Miglani and Miguel Martin and Edward Wang and Bilal Alsallakh and Jonathan Reynolds and Alexander Melnikov and Natalia Kliushkina and Carlos Araya and Siqi Yan and Orion Reblitz-Richardson},
      year={2020},
      eprint={2009.07896},
      archivePrefix={arXiv},
      primaryClass={cs.LG},
      url={https://arxiv.org/abs/2009.07896}, 
}

@misc{ribeiro2016whyitrustyou,
      title={"Why Should I Trust You?": Explaining the Predictions of Any Classifier}, 
      author={Marco Tulio Ribeiro and Sameer Singh and Carlos Guestrin},
      year={2016},
      eprint={1602.04938},
      archivePrefix={arXiv},
      primaryClass={cs.LG},
      url={https://arxiv.org/abs/1602.04938}, 
}

@misc{sundararajan2017axiomaticattributiondeepnetworks,
      title={Axiomatic Attribution for Deep Networks}, 
      author={Mukund Sundararajan and Ankur Taly and Qiqi Yan},
      year={2017},
      eprint={1703.01365},
      archivePrefix={arXiv},
      primaryClass={cs.LG},
      url={https://arxiv.org/abs/1703.01365}, 
}

@misc{shrikumar2019learningimportantfeaturespropagating,
      title={Learning Important Features Through Propagating Activation Differences}, 
      author={Avanti Shrikumar and Peyton Greenside and Anshul Kundaje},
      year={2019},
      eprint={1704.02685},
      archivePrefix={arXiv},
      primaryClass={cs.CV},
      url={https://arxiv.org/abs/1704.02685}, 
}

@article{Selvaraju_2019,
   title={Grad-CAM: Visual Explanations from Deep Networks via Gradient-Based Localization},
   volume={128},
   ISSN={1573-1405},
   url={http://dx.doi.org/10.1007/s11263-019-01228-7},
   DOI={10.1007/s11263-019-01228-7},
   number={2},
   journal={International Journal of Computer Vision},
   publisher={Springer Science and Business Media LLC},
   author={Selvaraju, Ramprasaath R. and Cogswell, Michael and Das, Abhishek and Vedantam, Ramakrishna and Parikh, Devi and Batra, Dhruv},
   year={2019},
   month=oct, pages={336–359} }

@misc{rw2019timm,
  author = {Ross Wightman},
  title = {PyTorch Image Models},
  year = {2019},
  publisher = {GitHub},
  journal = {GitHub repository},
  doi = {10.5281/zenodo.4414861},
  howpublished = {\url{https://github.com/rwightman/pytorch-image-models}}
}

@misc{torchvision2016,
    title        = {TorchVision: PyTorch's Computer Vision library},
    author       = {{T}orchVision maintainers and contributors},
    year         = 2016,
    journal      = {GitHub repository},
    publisher    = {GitHub},
    howpublished = {\url{https://github.com/pytorch/vision}}
}

@misc{imagenet-object-localization-challenge,
    author = {Addison Howard and Eunbyung Park and Wendy Kan},
    title = {ImageNet Object Localization Challenge},
    year = {2018},
    howpublished = {\url{https://kaggle.com/competitions/imagenet-object-localization-challenge}},
    note = {Kaggle}
}

@misc{szegedy2014intriguingpropertiesneuralnetworks,
      title={Intriguing properties of neural networks}, 
      author={Christian Szegedy and Wojciech Zaremba and Ilya Sutskever and Joan Bruna and Dumitru Erhan and Ian Goodfellow and Rob Fergus},
      year={2014},
      eprint={1312.6199},
      archivePrefix={arXiv},
      primaryClass={cs.CV},
      url={https://arxiv.org/abs/1312.6199}, 
}

@misc{mahendran2014understandingdeepimagerepresentations,
      title={Understanding Deep Image Representations by Inverting Them}, 
      author={Aravindh Mahendran and Andrea Vedaldi},
      year={2014},
      eprint={1412.0035},
      archivePrefix={arXiv},
      primaryClass={cs.CV},
      url={https://arxiv.org/abs/1412.0035}, 
}

@misc{bykov2021noisegrad,
      title={NoiseGrad: Enhancing Explanations by Introducing Stochasticity to Model Weights}, 
      author={Kirill Bykov and Anna Hedström and Shinichi Nakajima and Marina M. -C. Höhne},
      year={2022},
      eprint={2106.10185},
      archivePrefix={arXiv},
      primaryClass={cs.LG},
      url={https://arxiv.org/abs/2106.10185}, 
}

@misc{heo2019foolingneuralnetworkinterpretations,
      title={Fooling Neural Network Interpretations via Adversarial Model Manipulation}, 
      author={Juyeon Heo and Sunghwan Joo and Taesup Moon},
      year={2019},
      eprint={1902.02041},
      archivePrefix={arXiv},
      primaryClass={cs.LG},
      url={https://arxiv.org/abs/1902.02041}, 
}

@misc{dombrowski2019explanationsmanipulatedgeometryblame,
      title={Explanations can be manipulated and geometry is to blame}, 
      author={Ann-Kathrin Dombrowski and Maximilian Alber and Christopher J. Anders and Marcel Ackermann and Klaus-Robert Müller and Pan Kessel},
      year={2019},
      eprint={1906.07983},
      archivePrefix={arXiv},
      primaryClass={stat.ML},
      url={https://arxiv.org/abs/1906.07983}, 
}

@misc{wang2021pyramidvisiontransformerversatile,
      title={Pyramid Vision Transformer: A Versatile Backbone for Dense Prediction without Convolutions}, 
      author={Wenhai Wang and Enze Xie and Xiang Li and Deng-Ping Fan and Kaitao Song and Ding Liang and Tong Lu and Ping Luo and Ling Shao},
      year={2021},
      eprint={2102.12122},
      archivePrefix={arXiv},
      primaryClass={cs.CV},
      url={https://arxiv.org/abs/2102.12122}, 
}

@misc{wrobel2026davedistributionawareattributionvit,
      title={DAVE: Distribution-aware Attribution via ViT Gradient Decomposition}, 
      author={Adam Wróbel and Siddhartha Gairola and Jacek Tabor and Bernt Schiele and Bartosz Zieliński and Dawid Rymarczyk},
      year={2026},
      eprint={2602.06613},
      archivePrefix={arXiv},
      primaryClass={cs.CV},
      url={https://arxiv.org/abs/2602.06613}, 
}

@misc{ariasduart2022focusratingxaimethods,
      title={Focus! Rating XAI Methods and Finding Biases}, 
      author={Anna Arias-Duart and Ferran Parés and Dario Garcia-Gasulla and Victor Gimenez-Abalos},
      year={2022},
      eprint={2109.15035},
      archivePrefix={arXiv},
      primaryClass={cs.LG},
      url={https://arxiv.org/abs/2109.15035}, 
}
\bibliographystyle{tmlr}

\newpage
\appendix


\newpage

\section*{Appendix Contents}

\begingroup
\setcounter{tocdepth}{2}

\apptocsection{app:appendix_motivation}
\apptocsubsection{sec:appendix_hard_gating_example}
\apptocsubsection{sec:appendix_interference}
\apptocsubsection{sec:appendix_maxpool}
\apptocsubsection{sec:appendix_backward_only}

\apptocsection{app:bias_robustness}
\apptocsubsection{app:bias_terms}
\apptocsubsection{app:bias_direction_selection}
\apptocsubsection{app:bias_directional_interpretation}
\apptocsubsection{app:bias_gating_shifts}
\apptocsubsection{sec:normalization_bias_robustness}

\apptocsection{app:layer_pullbacks}
\apptocsubsection{sec:soft_adjoint_activations}
\apptocsubsection{sec:soft_adjoint_pool}
\apptocsubsection{sec:soft_adjoint_norm}
\apptocsubsection{sec:soft_adjoint_affine}
\apptocsubsection{sec:soft_adjoint_attention}

\apptocsection{app:appendix_pytorch}
\apptocsubsection{app:backward_primitives}
\apptocsubsection{app:model_versions}
\apptocsubsection{app:computing_pullbacks}
\apptocsubsection{app:reproducibility}

\apptocsection{app:path_expansion}

\apptocsection{app:metrics}
\apptocsection{app:hyperparameter_sensitivity}
\apptocsection{app:figures_and_tables}

\endgroup


\begin{table}[tb]
\caption{PVT results on 1000 ImageNet validation samples.
Formatting conventions follow Table~\ref{table:resnet50_results}.}
\label{table:pvt_results}
\begin{center}
\begin{tabular}{lrrrrrr}
\toprule
Method
& Infidelity $\downarrow$
& Focus $\uparrow$
& Faith.Corr $\uparrow$
& Mono.Corr $\uparrow$
& Max.Sens $\downarrow$
& Rand.Logit $\downarrow$ \\
\midrule
Grad
& $\phantom{0}8.914_{\pm 07.89}$ & $0.557_{\pm 0.18}$ & $0.105_{\pm 0.33}$ & $0.185_{\pm 0.43}$ & $1.034_{\pm 0.16}$ & $0.030_{\pm 0.30}$ \\
\textbf{SoftPb}
& $\phantom{0}6.264_{\pm 06.40}$ & $0.617_{\pm 0.18}$ & $0.119_{\pm 0.34}$ & $\deterioration{0.164_{\pm 0.43}}$ & $\deterioration{1.066_{\pm 0.18}}$ & $0.006_{\pm 0.35}$ \\
\pairsep
GA
& $\phantom{0}4.506_{\pm 04.07}$ & $0.551_{\pm 0.10}$ & $0.118_{\pm 0.33}$ & $0.164_{\pm 0.35}$ & $1.242_{\pm 0.07}$ & $0.006_{\pm 0.06}$ \\
\textbf{PA}
& $\phantom{0}\best{1.634}_{\pm 01.03}$ & $0.554_{\pm 0.10}$ & $\best{0.122}_{\inconclusive{\pm 0.36}}$ & $0.219_{\pm 0.33}$ & $0.855_{\pm 0.12}$ & $0.121_{\pm 0.11}$ \\
\pairsep
SmoothG
& $\phantom{0}8.798_{\pm 07.08}$ & $0.594_{\pm 0.11}$ & $0.096_{\pm 0.38}$ & $\best{0.280}_{\pm 0.38}$ & $0.569_{\pm 0.10}$ & $0.012_{\pm 0.13}$ \\
\textbf{SmoothPb}
& $\phantom{0}4.974_{\pm 05.46}$ & $\best{0.635}_{\pm 0.10}$ & $\inconclusive{0.102_{\pm 0.39}}$ & $\deterioration{0.231_{\pm 0.38}}$ & $\best{0.519}_{\pm 0.09}$ & $0.008_{\pm 0.23}$ \\
\pairsep
FusionG
& $\phantom{0}6.673_{\pm 05.60}$ & $0.606_{\pm 0.10}$ & $0.101_{\pm 0.39}$ & $0.262_{\pm 0.38}$ & $0.979_{\pm 0.16}$ & $0.001_{\pm 0.11}$ \\
\textbf{FusionPb}
& $\phantom{0}5.156_{\pm 05.72}$ & $0.614_{\pm 0.11}$ & $\inconclusive{0.093_{\pm 0.39}}$ & $\deterioration{0.200_{\pm 0.38}}$ & $0.753_{\pm 0.11}$ & $0.007_{\pm 0.19}$ \\
\midrule
IG
& $12.578_{\pm 08.39}$ & $0.571_{\pm 0.19}$ & $0.068_{\pm 0.32}$ & $0.170_{\pm 0.45}$ & $1.374_{\pm 0.37}$ & $0.008_{\pm 0.33}$ \\
GradShap
& $12.433_{\pm 08.45}$ & $0.562_{\pm 0.19}$ & $0.067_{\pm 0.32}$ & $0.169_{\pm 0.45}$ & $2.471_{\pm 1.56}$ & $0.003_{\pm 0.34}$ \\
DeepLift
& $13.132_{\pm 08.83}$ & $0.548_{\pm 0.21}$ & $0.085_{\pm 0.31}$ & $0.186_{\pm 0.45}$ & $1.047_{\pm 0.15}$ & $0.029_{\pm 0.32}$ \\
\bottomrule
\end{tabular}
\end{center}
\end{table}

\begin{table}[tb]
\caption{VGG results on 1000 ImageNet validation samples.
Formatting conventions follow Table~\ref{table:resnet50_results}.}
\label{table:vgg_results}
\begin{center}
\begin{tabular}{lrrrrrr}
\toprule
Method
& Infidelity $\downarrow$
& Focus $\uparrow$
& Faith.Corr $\uparrow$
& Mono.Corr $\uparrow$
& Max.Sens $\downarrow$
& Rand.Logit $\downarrow$ \\
\midrule
Grad
& $100.090_{\pm 107.94}$ & $0.544_{\pm 0.13}$ & $0.457_{\pm 0.35}$ & $0.316_{\pm 0.37}$ & $0.898_{\pm 0.14}$ & $0.053_{\pm 0.29}$ \\
\textbf{SoftPb}
& $16.303_{\pm 31.68}$ & $0.614_{\pm 0.11}$ & $\best{0.606}_{\pm 0.30}$ & $0.375_{\pm 0.39}$ & $\best{0.212}_{\pm 0.08}$ & $0.064_{\pm 0.33}$ \\
\pairsep
GA
& $18.523_{\pm 35.24}$ & $0.557_{\pm 0.10}$ & $0.508_{\pm 0.30}$ & $0.234_{\pm 0.32}$ & $1.217_{\pm 0.06}$ & $0.001_{\pm 0.07}$ \\
\textbf{PA}
& $\phantom{0}\best{5.616}_{\pm 13.18}$ & $0.623_{\pm 0.11}$ & $0.570_{\pm 0.29}$ & $0.312_{\pm 0.37}$ & $0.390_{\pm 0.11}$ & $0.136_{\pm 0.13}$ \\
\pairsep
SmoothG
& $52.044_{\pm 70.80}$ & $0.562_{\pm 0.11}$ & $0.513_{\pm 0.34}$ & $\best{0.385}_{\pm 0.36}$ & $0.781_{\pm 0.11}$ & $0.025_{\pm 0.19}$ \\
\textbf{SmoothPb}
& $14.030_{\pm 30.57}$ & $\best{0.625}_{\pm 0.12}$ & $0.545_{\pm 0.32}$ & $\deterioration{0.309_{\pm 0.40}}$ & $0.398_{\pm 0.11}$ & $0.048_{\pm 0.32}$ \\
\pairsep
FusionG
& $38.651_{\pm 60.23}$ & $0.561_{\pm 0.11}$ & $0.506_{\pm 0.34}$ & $0.363_{\pm 0.34}$ & $0.955_{\pm 0.13}$ & $0.013_{\pm 0.16}$ \\
\textbf{FusionPb}
& $16.056_{\pm 34.46}$ & $0.614_{\pm 0.12}$ & $\inconclusive{0.513_{\pm 0.33}}$ & $\deterioration{0.315_{\pm 0.39}}$ & $0.513_{\pm 0.11}$ & $0.035_{\pm 0.29}$ \\
\midrule
IG
& $75.796_{\pm 92.29}$ & $0.547_{\pm 0.16}$ & $0.434_{\pm 0.36}$ & $0.379_{\pm 0.41}$ & $0.752_{\pm 0.15}$ & $0.046_{\pm 0.27}$ \\
GradShap
& $76.050_{\pm 94.76}$ & $0.545_{\pm 0.16}$ & $0.430_{\pm 0.36}$ & $0.381_{\pm 0.41}$ & $1.086_{\pm 0.22}$ & $0.042_{\pm 0.26}$ \\
DeepLift
& $85.658_{\pm 97.55}$ & $0.535_{\pm 0.17}$ & $0.356_{\pm 0.37}$ & $0.358_{\pm 0.42}$ & $0.929_{\pm 0.15}$ & $0.057_{\pm 0.29}$ \\
\bottomrule
\end{tabular}
\end{center}
\end{table}

\section{Additional Motivation and Toy Examples}
\label{app:appendix_motivation}

This appendix provides supplementary mechanistic motivation for backward-only softening in hard-gated layers, particularly ReLU and MaxPool.
As discussed in the main text, a coherent feature may be only partially expressed across its supporting units.
Hard gating can aggravate this issue by completely removing negative components or routing the backward signal through a single winning location.
The examples below illustrate these effects and explain how softened backward gates can retain locally suppressed components without changing the pretrained model's forward computation.

\subsection{Hard gating can drop coherent components}\label{sec:appendix_hard_gating_example}

Consider a one-layer network with two directions $\vw_1,\vw_2\in\R^d$ and score
\begin{equation}\label{eq:appendix_relu_sum}
s(\vx) = \mathrm{ReLU}(\vw_1^\top \vx) + \mathrm{ReLU}(\vw_2^\top \vx)
\end{equation}
If $\vw_1^\top \vx>0$ but $\vw_2^\top \vx<0$, then the standard pullback is $\nabla_{\vx} s(\vx)=\vw_1$ and the $\vw_2$ component is completely
eliminated, regardless of whether $\vw_2$ corresponds to a semantically meaningful part of a larger feature.
A backward-only soft gate $\Phi((\vw_2^\top \vx)/\tau)$ yields a softened pullback
\begin{equation}
\tilde{\nu}(\vx) = \Phi\!\left(\frac{\vw_1^\top \vx}{\tau}\right)\vw_1 + \Phi\!\left(\frac{\vw_2^\top \vx}{\tau}\right)\vw_2
\end{equation}
which retains information about near-threshold directions and reduces discontinuities at region boundaries.

\subsection{Local relevance and destructive interference}\label{sec:appendix_interference}

Hard gating can also create locally inconsistent explanations when multiple directions overlap in input space.
Consider
\begin{equation}\label{eq:appendix_signed_sum}
s(\vx) = \mathrm{ReLU}(\vw_1^\top \vx) - \mathrm{ReLU}(\vw_2^\top \vx)
\end{equation}
with $\vw_2 = \vm\odot \vw_1$ where $\vm\in\{0,1\}^d$ is a binary mask that zeroes a large fraction of coordinates.
Suppose $\vw_1^\top \vx \gg 0$ and $\vw_2^\top \vx > 0$ but small, so that $\nabla_{\vx} s(\vx)=\vw_1-\vw_2=(\mathbf{1}-\vm)\odot \vw_1$.
Now consider a nearby point $\vx'$ such that $\vw_1^\top \vx'\gg 0$ but $\vw_2^\top \vx'<0$ (e.g., due to a sign flip in the masked
coordinates). Then $\nabla_{\vx} s(\vx')=\vw_1$.
Thus, the pullback changes abruptly across neighbouring inputs even when the dominant direction $\vw_1$ is stable.
This illustrates how steep gating can cause gradient-based explanations to depend on brittle activation patterns rather than on the underlying coherent direction.

\subsection{Hard routing can discard near-winning components}
\label{sec:appendix_maxpool}

A related effect occurs in MaxPool.
Consider
\[
s(\vx)=\max\{\vw_1^\top \vx,\vw_2^\top \vx\}.
\]
Away from ties, its standard pullback equals $\vw_1$ or $\vw_2$, depending on which response is larger.
When the responses are close, a small input perturbation can switch the entire backward signal between these directions.

If both responses correspond to related feature components, exclusive routing can hide information carried by the near-winning response.
A softened backward routing rule instead produces
\[
\tilde{\nu}(\vx)=p_1(\vx)\vw_1+p_2(\vx)\vw_2,
\qquad
p_i(\vx)\geq 0,
\qquad
p_1(\vx)+p_2(\vx)=1,
\]
where the weights vary smoothly with the responses.
Under a local perturbation model, the expected hard-routing weights are precisely the probabilities of each response winning.
This provides an interpretation of soft routing as retaining components that may become selected in a neighborhood of the input.

\subsection{Why backward-only softening differs from changing the forward nonlinearity}
\label{sec:appendix_backward_only}

Replacing ReLU or MaxPool with a smooth forward operation changes the function implemented by the pretrained network, including downstream activations and predictions.
Backward-only softening instead preserves the original forward state and modifies only the transport of the explanatory signal.

This separation also allows the backward softening scale to be chosen independently of the forward computation.
A smooth forward gate can still strongly suppress negative or non-winning components when its transition is sharp relative to the relevant local variation.
Backward-only softening directly controls this suppression during explanation.
However, components far outside the chosen softening scale can still receive negligible weight; the mechanism primarily recovers directions near activation or routing boundaries.


\section{Robustness to Biases}
\label{app:bias_robustness}

This appendix clarifies why the proposed gate-based interpretation is naturally robust to affine \emph{bias} terms, and how biases relate to neural network ``directional preferences'' (i.e., which input-space directions are locally amplified or suppressed).


\subsection{Handling Bias Terms via Matrix Augmentation}\label{app:bias_terms}

This appendix records a convenient bias-separation trick used for formal analysis.
The main method operates on input-space vectors; it is often useful to separate weight contributions from bias contributions.

Consider an affine map $\vy = \mW\vx + \vb$ with $\mW\in\R^{m\times d}$ and $\vb\in\R^m$.
Define augmented vectors and matrix:
\begin{equation}
\hat{\vx} :=
\begin{bmatrix} \vx \\ 1 \end{bmatrix}\in\R^{d+1},
\qquad
\hat{\mW} :=
\begin{bmatrix}
\mW & \vb \\
0 & 1
\end{bmatrix}
\in\R^{(m+1)\times(d+1)}
\end{equation}
Then $\hat{\vy}=\hat{\mW}\hat{\vx}$ expresses the affine map as a matrix multiplication in augmented coordinates.

Given a vector $\hat{\vu}\in\R^{m+1}$ and $\vu=\mathrm{proj}_{\R^m}(\hat{\vu})$, the standard adjoint is
\begin{equation}
\hat{\mW}^\top \hat{\vu}
=
\begin{bmatrix}
\mW^\top \vu \\
\vb^\top \vu  + \evu_{m+1}
\end{bmatrix}
\end{equation}
so the last coordinate accumulates bias-like contributions.
Note that
\begin{equation}\label{eq:bias_residual_closed_form}
\langle \hat{\vu}, \hat{\mW}\hat{\vx}\rangle =
\langle \hat{\mW}^\top\hat{\vu}, \hat{\vx}\rangle =
\langle \mW^\top \vu, \vx\rangle + (\vb^\top \vu + \evu_{m+1})
\end{equation}
i.e. the scalar dot product $\langle \mW^\top \vu, \vx\rangle$ involving the linear component $\mW^\top$ of the adjoint $\hat{\mW}^\top$ approximates $\langle \hat{\vu}, \hat{\mW}\hat{\vx}\rangle$ up to the bias term $(\vb^\top \vu + \evu_{m+1})$.



\subsection{Biases, pre-activations, and local direction selection}\label{app:bias_direction_selection}

Consider a standard feed-forward layer with an element-wise activation,
\begin{equation}
\label{eq:affine_layer}
\vz_\ell = \mW_\ell\, \vx_{\ell-1} + \vb_\ell,
\qquad
\vx_\ell = \phi\!\left(\vz_\ell\right)
\end{equation}
where $\phi$ is an elementwise nonlinearity (e.g., ReLU, SiLU, GELU).
A convenient view for a broad class of activations is that the nonlinearity can be written as an \emph{input-dependent gating} (or modulation) applied to the pre-activation:
\begin{equation}
\label{eq:gate_form}
\vx_\ell = g\!\left(\vz_\ell\right) \odot \vz_\ell
\end{equation}
where $g(\cdot)$ is applied elementwise and produces values in $[0,1]$ (or in a bounded range).
For ReLU, one may take $g(\vz)=\1{\vz>0}$ (hard gating) or a softened surrogate gate (e.g., a sigmoid with temperature).
For smooth activations such as SiLU/GELU, $g(\vz)$ is naturally soft and varies continuously with $\vz$.

Equation~\ref{eq:gate_form} emphasizes the key point: \emph{activations implement a local selection of directions}.
The network amplifies components aligned with directions that induce high gate values and suppresses other components that induce low gate values.
Thus, the model’s inductive bias can be understood through the geometry of these gates: which directions in input space, under the current data point, are treated as ``important'' (high gating) versus ``discarded'' (low gating).

\subsection{Why biases are not central to the directional interpretation}\label{app:bias_directional_interpretation}

Biases shift pre-activations $\vz_\ell$ by an additive term $\vb_\ell$.
This shift can move units between regimes (e.g., negative to positive in ReLU), thereby changing which gates are active.
However, from a gate-centric viewpoint, what matters for interpretation is not the absolute value of $\vz_\ell$ per se, but rather the \emph{local gating state} $g(\vz_\ell)$ that determines which directions are selected.

In particular, the same directional selection behavior can be realized in two conceptually different ways:
\begin{enumerate}
    \item \textbf{Explicit-bias gating:} compute $\vz_\ell=\mW_\ell\, \vx_{\ell-1}+\vb$ and then gate using $g(\vz_\ell)$.
    \item \textbf{Virtual-bias gating:} keep a purely linear signal $\tilde{\vz}_\ell=\mW_\ell\, \vx_{\ell-1}$, but compute gates \emph{as if} the bias were present, i.e.,
    \begin{equation}
    \label{eq:virtual_bias_gating}
    \tilde{\vx}_\ell \;=\; g(\tilde{\vz}_\ell+\vb_\ell) \odot \tilde{\vz}_\ell
    \end{equation}
\end{enumerate}

The second construction highlights a robustness intuition: biases primarily act by shifting which gates are selected, not by contributing a direction of their own.
If the goal is to represent the network’s action in terms of input-space directions, then the quantities to ``pull back'' are the gate values (direction-selection signals), not the bias offsets.

\subsection{Negative regimes and bias-induced gating shifts}\label{app:bias_gating_shifts}

Bias terms can be particularly salient when a unit’s dot products $\mW\vx$ lie largely in the negative regime.
For hard-threshold activations (e.g., ReLU), if $\mW\vx$ is negative for many samples, then an appropriately chosen bias can flip the unit into the active regime.
This corresponds to an effective change in the gating domain: the decision boundary for activation is translated.
Thus, when bias shifts move units across gating regimes, the pullback directional structure updates accordingly.
The interpretation remains stable in the sense that it continues to attribute importance to the \emph{input directions} that are locally selected after the shift, rather than to the additive offset itself.

\subsection{Connection to normalization layers}\label{sec:normalization_bias_robustness}

This perspective is also consistent with normalization layers.
Batch Normalization and related methods reduce sensitivity to shifts in activation statistics and can effectively diminish the functional role of explicit biases (often making biases redundant in preceding layers).
From a gate-centric viewpoint, normalization stabilizes the distribution of pre-activations, thereby stabilizing gate selection and reducing brittle regime switching.
In other words, normalization improves robustness partly by making the direction-selection mechanism (gating) less sensitive to nuisance offsets.
Although normalization layers introduce their own bias through mean subtraction, this effect can be treated analogously to Equation~\ref{eq:virtual_bias_gating} by using a virtual normalization step solely to compute the appropriate gating.





\section{Layer-wise Soft Adjoints}\label{app:layer_pullbacks}

This appendix specifies the layer-wise \emph{soft adjoint} operators used to instantiate the modified backward
propagation. Throughout, a layer $h_\ell$ produces activations
$\vx_\ell = h_\ell(\vx_{\ell-1})$ and standard backprop propagates cotangents via
$\vv_{\ell-1} = \mathrm{VJP}_\ell(\vx_{\ell-1};\vv_\ell)=\mJ_\ell(\vx_{\ell-1})^\top \vv_\ell$.
When $h_\ell$ is an elementwise activation function, we denote its input by
$\vz_\ell := \vx_{\ell-1}$ for notational convenience.
Our soft adjoint replaces $\mathrm{VJP}_\ell$ with $\tilde{\mathrm{VJP}}_\ell$ while keeping the \emph{forward pass unchanged}.

\paragraph{Design principle}
For layers that contain \emph{hard routing} or \emph{steep gating} (e.g., ReLU masks, max-pooling switches,
sharp attention), we replace the backward gating weights with a \emph{softer} function controlled by a temperature
$\tau>0$. For affine layers, we keep the standard VJP.
For Normalization and Self-Attention layers, we block the backward flow through the
statistics and softmax computations, respectively, with the option to soften the softmax by temperature $\tau > 0$.

\subsection{Activation layers (ReLU, SiLU, GELU)}\label{sec:soft_adjoint_activations}

Consider an element-wise activation $\vx_\ell=\phi(\vz_\ell)$ applied to pre-activations $\vz_\ell$. Given a backward signal $\vv_\ell$ with respect to $\vx_\ell$, standard backprop through this activation layer gives
\begin{equation}
\vv_{\ell - 1} \;=\; \vv_\ell \odot \phi'(\vz_\ell)
\end{equation}
Assume that $\phi(\vz) = \vz\odot g(\vz)$ for some \emph{gating function} $g$.
We define a \emph{soft gate} $\tilde{g}_\phi(\vz_\ell;\tau)$ and use
\begin{equation}\label{eq:soft_activation_vjp}
\tilde{\mathrm{VJP}}_{\phi}(\vz_\ell;\vv_\ell;\tau)
\;:=\;
\vv_\ell \odot \tilde{g}_\phi(\vz_\ell;\tau)
\end{equation}

\paragraph{ReLU}
For $\phi(\vz)= \vz\odot\mathbf{1}[\vz>0]$ we soften the gate with a Normal CDF function $\Phi$:
\begin{equation}\label{eq:soft_relu_gate}
\tilde{g}_{\mathrm{ReLU}}(\vz;\tau)=\Phi\!\left(\frac{\vz}{\tau}\right)
\end{equation}
so that negative but near-threshold units contribute non-trivially in the backward pass.

\paragraph{SiLU (Swish)}
For $\phi(\vz)=\vz\odot\sigma(\vz)$, standard backprop uses
$\phi'(\vz)=\sigma(\vz)+\vz\odot\sigma(\vz)\odot(1-\sigma(\vz))$, i.e. the exact derivative.
Our adjoint view instead transports the signal through the dynamic linear factor $\vz\mapsto \sigma(\vz)\odot\vz$, without differentiating through the gate.
We therefore use the temperature-scaled gate
\begin{equation}
\tilde{g}_{\mathrm{SiLU}}(\vz;\tau)=\sigma\!\left(\frac{\vz}{\tau}\right)
\end{equation}


\paragraph{GELU}
For $\phi(\vz) = \vz\odot\Phi(\vz)$, we analogously use a temperature-scaled gates:
\begin{equation}\label{eq:soft_gelu_gate}
\tilde{g}_{\mathrm{GELU}}(\vz;\tau)=\Phi\!\left(\frac{\vz}{\tau}\right)
\end{equation}
where $\Phi$ is the Normal CDF.

\paragraph{Remark (hyperparameter meaning)}
Larger $\tau$ makes the gate \emph{less steep} in the backward pass.

\subsection{Max pooling}\label{sec:soft_adjoint_pool}

Max pooling performs hard routing in the forward pass and routes gradients to the argmax index in standard backprop.
For a pooling window with entries $\vz = \{\evz_i\}_{i=1}^m$, standard backward assigns all mass to $i^\star=\arg\max_i \evz_i$.

We define a backward-only \emph{soft routing distribution}
\begin{equation}\label{eq:pool_soft_routing}
p_i(\vz;\tau) \;=\; \frac{\exp(\evz_i/\tau)}{\sum_{j=1}^m \exp(\evz_j/\tau)}
\end{equation}
and distribute the incoming cotangent signal $v\in\R$ across the window:
\begin{equation}\label{eq:soft_pool_vjp}
\tilde{\mathrm{VJP}}_{\mathrm{MaxPool}}(\vz;v;\tau)_i
\;:=\;
p_i(\vz;\tau)\,v
\end{equation}
The forward max operation remains unchanged; only the backward routing is softened. We accumulate the signals from overlapping windows.

\subsection{Normalization layers (BatchNorm, LayerNorm)}\label{sec:soft_adjoint_norm}

We treat normalization layers as non-gating operations and therefore do not apply softening.
BatchNorm reduces to an affine transformation at inference time.
For LayerNorm, we stop gradients through the mean/variance computation and backpropagate using the cached normalization statistics.

\subsection{Affine layers (Linear, Conv2d)}\label{sec:soft_adjoint_affine}

For affine layers, we use the standard VJP.
If $\vy=\mW\vx$ then $\mathrm{VJP}(\vx;\vv)=\mW^\top \vv$.
For convolutions, the VJP is the transposed convolution with the same parameters (stride, padding, groups, etc.).
Thus,
\begin{equation}
\tilde{\mathrm{VJP}}_{\mathrm{Affine}} \equiv \mathrm{VJP}_{\mathrm{Affine}}
\end{equation}


\subsection{Self-Attention}\label{sec:soft_adjoint_attention}

We describe a single-head attention block; multi-head attention is applied per head and concatenated.
Let $\mX\in\R^{n\times d}$ be token embeddings, and define
$\mQ=\mX\mW_Q,\; \mK=\mX\mW_K,\; \mV=\mX\mW_V$.
Let attention logits be
\begin{equation}
\mM \;=\; \frac{\mQ\mK^\top}{\sqrt{d_k}} \in \R^{n\times n}
\end{equation}
and attention weights $\mA=\mathrm{softmax}(\mM)$ (row-wise). The attention output is $\mY=\mA\mV$.

\paragraph{Blocked backward through softmax}
We keep the forward attention weights $\mA$ (and thus the forward output) unchanged.
In the backward pass, we block gradient flow through the softmax and through the computation of $\mQ$ and $\mK$,
so that the signal propagates only through the Value branch.

\paragraph{Soft backward attention (optional)}
We optionally use a temperature-softened attention in the backward pass only
\begin{equation}\label{eq:soft_attention_map}
\tilde{\mA} \;=\; \mathrm{softmax}\!\left(\frac{\mM}{\tau}\right)
\end{equation}
and propagate the signal through $\mY=\tilde{\mA}\mV$ while still blocking gradients through $\mQ$ and $\mK$.
Larger $\tau$ yields a less steep attention map.



\section{PyTorch Implementation Details}\label{app:appendix_pytorch}

We implement semantic pullbacks in PyTorch by (i) running a standard forward pass and (ii) computing soft pullbacks
using \emph{backward-only} modified VJPs.

\subsection{Backward-only modification via autograd primitives}\label{app:backward_primitives}

Our implementation relies on lightweight modifications of the backward pass while leaving the forward computation unchanged.
Concretely, we combine three generic mechanisms:

\begin{itemize}
    \item \textbf{Forward Gradient Injection (FGI)}~\citep{otte2024flexibleefficientsurrogategradient}
    for element-wise activation functions;
    \item \textbf{Strike-through trick}, which replaces a term in the backward pass by a user-defined surrogate;
    while keeping the original factor in the forward;
    \item \textbf{Detach-based blocking} of gradient flow through selected computational paths.
\end{itemize}

\paragraph{Element-wise activations}
For ReLU, GELU, and SiLU, we use Forward Gradient Injection:
the forward computes the original activation $\vy=\phi(\vz)$, while the backward returns
\[
\mathrm{d}\vz = \mathrm{d}\vy \odot \tilde{g}(\vz)
\]
where $\tilde{g}$ is the softened gate.

\paragraph{Strike-through trick}
Given a forward expression $\vy = \vb$, we implement
\[
\vy = \va - \mathrm{stopgrad}(\va) + \mathrm{stopgrad}(\vb)
\]
so that the forward value is unchanged, but the backward uses gradients of $\va$ instead of $\vb$.
We use this pattern to substitute hard routing terms by their softened counterparts.

\paragraph{Blocking gradient flow}
In several layers we explicitly stop gradients through selected paths using \texttt{detach}.
Examples include blocking gradients through normalization statistics and the computation
of routing weights in pooling or attention layers, while still allowing gradients to flow through the Value branch in the latter.

\subsection{Replacing modules in pretrained models}\label{app:model_versions}

\paragraph{Architectures}
We use pretrained models from \texttt{torchvision}~\citep{torchvision2016} and \texttt{timm}~\citep{rw2019timm}.
Specifically, we evaluate \texttt{resnet50} and \texttt{vgg11\_bn} from \texttt{torchvision}
and \texttt{pvt\_v2\_b1} from the \texttt{timm} library, all with \texttt{pretrained=True} flag.
For these models, we traverse the module tree and replace selected layers:

\begin{itemize}
    \item \texttt{nn.ReLU}, \texttt{nn.GELU}, \texttt{nn.SiLU} $\rightarrow$ FGI-based wrappers;
    \item \texttt{nn.MaxPool2d} $\rightarrow$ custom pooling layer with (detached) soft routing;
    \item \texttt{nn.LayerNorm} $\rightarrow$ wrapper that blocks gradient flow through normalization statistics;
    \item \texttt{pvt\_v2.Attention} layer $\rightarrow$ explicit attention implementations exposing logits $\mM$.
\end{itemize}

All replacements preserve the exact forward outputs of the original model.

\subsection{Computing Soft Pullbacks and Pullback Ascent}\label{app:computing_pullbacks}

Given an input tensor $\vx$ with \texttt{requires\_grad=True} and a target class $c$, we compute the \emph{soft pullback}
by backpropagating from the scalar score $s_{\ve_c}(\vx)$:
\begin{itemize}
    \item run a forward pass to obtain logits $f(\vx)$;
    \item set $s = f(\vx)[c]$ (or a batch sum);
    \item call \texttt{torch.autograd.grad(s, x)} to obtain the soft pullback in input space.
\end{itemize}
Because the model modules have been replaced by backward-only softened wrappers, this gradient corresponds to $\tilde{\nu}_{\ve_c}(\vx)$ from Equation~\ref{eq:soft_pullback}.
The PullbackAscent is computed according to Algorithm~\ref{alg:pullback_ascent}.



\begin{algorithm}[t]
\caption{PullbackAscent (PA)}
\label{alg:pullback_ascent}
\begin{algorithmic}
\STATE {\bfseries Input:} input $\vx$, target class $c$, steps $K$, step size $\alpha$
\STATE {\bfseries Output:} class-conditioned explanation $\tilde{\nu}_{\ve_c}^*(\vx)$
\STATE $\vx^{(0)} \leftarrow \vx$
\FOR{$t=0$ {\bfseries to} $K-1$}
    \STATE Compute soft pullback $\tilde{\nu}_{\ve_c}(\vx^{(t)})$
    \STATE Optionally normalize $\tilde{\nu}_{\ve_c}(\vx^{(t)})$ (unit $\ell_2$)
    \STATE $\vx^{(t+1)} \leftarrow \vx^{(t)} + \alpha\,\tilde{\nu}_{\ve_c}(\vx^{(t)})$
\ENDFOR
\STATE Optionally clip $\vx^{(K)}$ to the valid pixel range
\STATE $\tilde{\nu}_{\ve_c}^*(\vx) \leftarrow \vx^{(K)} - \vx$
\end{algorithmic}
\end{algorithm}

\subsection{Reproducibility notes}\label{app:reproducibility}

We fix random seeds, run models in \texttt{eval()} mode, and enable deterministic PyTorch operations when available.
Note that gradient computation on CUDA may still be nondeterministic for some operators.


\section{Path perspective}\label{app:path_expansion}

Let $\mW_\ell \in \R^{d_\ell \times d_{\ell-1}}$ for $\ell=1,\dots,L$,
$\vx \in \R^{d_0}$ and $\vu \in \R^{d_L}$. We define
\begin{equation}
s_{\vu}(\vx) \;:=\; \vu^\top \mW_L \mW_{L-1}\cdots \mW_1 \vx
\end{equation}

Expanding the matrix product yields a sum over all paths
$\pi = (i_0,i_1,\dots,i_L)$ with $i_\ell \in [d_\ell]$:
\begin{equation}
s_{\vu}(\vx)
=
\sum_{i_L=1}^{d_L}
\sum_{i_{L-1}=1}^{d_{L-1}}
\cdots
\sum_{i_0=1}^{d_0}
\evu_{i_L}
\left(
\prod_{\ell=1}^{L} (\mW_\ell)_{i_\ell,i_{\ell-1}}
\right)
\evx_{i_0}
\end{equation}

Equivalently, letting
\begin{equation}
\mathcal{P} := [d_0] \times [d_1] \times \cdots \times [d_L]
\end{equation}
we may write
\begin{equation}
s_{\vu}(\vx)
=
\sum_{\pi=(i_0,\dots,i_L)\in\mathcal{P}}
\evu_{i_L}
\prod_{\ell=1}^{L} (\mW_\ell)_{i_\ell,i_{\ell-1}}
\evx_{i_0}
\end{equation}

Assume now that after each linear layer $\mW_\ell$ we insert an element-wise activation, represented by a diagonal gating matrix $\mG_\ell(\vx)$.
Thus the network alternates locally as
\begin{equation}
\vx_\ell = \mG_\ell(\vx) \mW_\ell \vx_{\ell-1}
\end{equation}

where $\mG_\ell(\vx)$ is diagonal and its entries depend on the corresponding pre-activation $\vz_\ell(\vx)=\mW_\ell \vx_{\ell-1}$.
For example, $\mG_\ell(\vx)=\mathrm{diag}(\sigma(\vz_\ell(\vx)))$ for SiLU, while
$\mG_\ell(\vx)=\mathrm{diag}(\mathbf{1}\{\vz_\ell(\vx)>0\})$ for ReLU.

With this notation, the path-wise expansion naturally factorizes.
Each path contribution decomposes into two components: a \emph{weight component} and a \emph{gating component}.
Explicitly, for a path $\pi = (i_0,\dots,i_L)$ we may write
\begin{equation}\label{eq:path_decomposition}
s_{\vu}(\vx)
=
\sum_{\pi\in\mathcal{P}}
\evu_{i_L}
\underbrace{
\prod_{\ell \in \mathcal{L}_W} (\mW_\ell)_{i_\ell,i_{\ell-1}}
}_{\text{weight of the path}}
\;
\underbrace{
\prod_{\ell \in \mathcal{L}_G} (\mG_\ell(\vx))_{i_\ell,i_\ell}
}_{\text{gating of the path}}
\;
\evx_{i_0}
\end{equation}
where $\mathcal{L}_W$ and $\mathcal{L}_G$ denote the sets of weight and gating layers, respectively.

This path-based perspective was first systematically studied in~\citet{lakshminarayanan2021neuralpathfeaturesneural}.
For purely feedforward or convolutional networks, the weight component is independent of the input $\vx$,
while \textbf{all input-dependence is captured by the gating component}.
Consequently, the dynamically linear nature of such networks arises entirely from the gating structure, i.e. the network is linear under the path assignment.

\textbf{Softening the pullback explicitly smooths the gating component}.
As a result, it relatively increases the contribution of \textbf{\emph{strong paths}} in Equation~\ref{eq:path_decomposition},
i.e.\ paths that predominantly traverse neurons with high activation (strong gating).
Our empirical findings, demonstrating both quantitative and qualitative advantages of Semantic Pullbacks,
strongly suggest that the network organizes an important part of its computation along these salient paths.

This perspective offers a potential mechanistic view of both network computation and training dynamics,
and may further inspire principled pruning or compression algorithms. In particular, we hypothesize that the strong paths mostly get fixed early on during training, i.e. the network learning dynamics can be approximated by a noisy kernel machine from a certain early point in time.
We leave a systematic investigation of these directions for future work.


\section{Quantus evaluation metrics}
\label{app:metrics}

Let $s_\vu:\R^d \to \R$ denote the scalar score selected by $\vu$ and let $\vx\in\R^d$ be an input image flattened to $d$ pixel coordinates.
An attribution method produces an explanation $e(\vx;\vu)\in\R^d$; we write $e(\vx)$ for brevity.
Let $\vx_0\in\R^d$ be a perturbation baseline and let $\odot$ denote element-wise multiplication.
For a binary mask $\vm\in\{0,1\}^d$, we write the corresponding perturbed input as
\begin{equation}
\vx^{(\vm)} = (\mathbf{1}-\vm)\odot\vx + \vm\odot\vx_0
\end{equation}
For consistency across metrics, we renormalize images to the range $[-1,1]$ before evaluation and disable Quantus' default channel reduction by setting \texttt{reduce\_axes=()}, so that explanations are evaluated in the full input space.

\paragraph{Infidelity (Faithfulness)}
Infidelity formalizes agreement between the explanation and the actual effect of input perturbations~\citep{yeh2019infidelitysensitivityexplanations}.
Let $I\in\R^d$ be a random perturbation, e.g., masking noise, Gaussian noise, or structured occlusion.
The infidelity of $e(\vx)$ is
\begin{equation}
\mathrm{Infid}(\vx)
=
\mathbb{E}_{I}
\Big[
\big(
\langle e(\vx), I\rangle
-
(s_\vu(\vx)-s_\vu(\vx-I))
\big)^2
\Big]
\end{equation}
Lower Infidelity is better: the explanation should predict the score change induced by the perturbation.

We place particular emphasis on Infidelity, since it explicitly evaluates whether an explanation captures the correct \emph{directional} effect of perturbations.
This is especially appropriate for image data, where pixels are strongly correlated into edges, textures, and object parts, and where meaningful changes are rarely independent single-pixel interventions.
In our Quantus setup, $I$ is implemented via baseline-replacement of $K$ square patches; we use uniform perturbations in $[-1,1]$ and apply optimal rescaling of the attribution map before evaluation (consistent with the original paper), so that the metric reflects directional agreement rather than arbitrary attribution scale.

\paragraph{Faithfulness Correlation (Faithfulness)}
Faithfulness Correlation measures whether regions assigned higher relevance produce larger score changes when perturbed~\citep{ijcai2020p0417}.
Given masks $\{\vm_k\}_{k=1}^K$, for each mask we compute
\begin{align}
b_k &= \langle e(\vx), \vm_k\rangle, \\
\Delta_k &= s_\vu(\vx)-s_\vu\!\big(\vx^{(\vm_k)}\big)
\end{align}
The metric is the Pearson correlation
\begin{equation}
\mathrm{FaithCorr}(\vx)
=
\mathrm{corr}\big(\{b_k\}_{k=1}^K,\{\Delta_k\}_{k=1}^K\big)
\end{equation}
Higher Faithfulness Correlation is better.

In its default form, this metric is less well matched to image data because random individual-pixel subsets treat pixels as independent variables.
We therefore modify the Quantus setup to use patch-based perturbations instead of random pixel subsets.
This evaluates faithfulness under spatially coherent image perturbations, matching the scale of object parts and visual evidence in natural images.
Since any perturbation distribution induces a corresponding evaluation bias, we interpret these scores as faithfulness under this image-specific protocol rather than as distribution-independent guarantees.
We also set \texttt{abs=True}, ranking features by attribution magnitude, and use \texttt{perturb\_baseline=0}, which corresponds to gray-value replacement after normalization to $[-1,1]$.



\paragraph{Monotonicity Correlation (Faithfulness)}
Quantus implements Monotonicity Correlation as the correlation between attribution mass and prediction variance under perturbations~\citep{nguyen2020quantitativeaspectsmodelinterpretability}.
Let $\pi$ be an ordering of coordinates such that $|e(\vx)_{\pi(1)}|\le \dots \le |e(\vx)_{\pi(d)}|$.
For step size $h$, define $S_t=\{\pi((t-1)h+1),\dots,\pi(th)\}$.
For each step, perturb $\vx$ on $S_t$ repeatedly $R$ times and compute
\begin{equation}
v_t
=
\frac{1}{R}\sum_{r=1}^R
\big(s_\vu(\vx^{(S_t,r)})-s_\vu(\vx)\big)^2
\cdot \mathrm{inv\_pred},
\qquad
b_t
=
\sum_{i\in S_t} |e(\vx)_i|
\end{equation}
where $\mathrm{inv\_pred}=1/\max(|s_\vu(\vx)|,\varepsilon)^2$ is a stabilizing factor.
The metric is the Spearman correlation
\begin{equation}
\mathrm{MonoCorr}(\vx)
=
\rho_{\mathrm{Spearman}}\big(\{b_t\},\{v_t\}\big)
\end{equation}
Higher Monotonicity Correlation is better.

This metric partially accounts for image structure through the attribution-induced ordering, but its perturbation model remains less directly aligned with visual feature directions than Infidelity.
We therefore interpret it as a complementary faithfulness score.
In our setup, we use uniform perturbations in $[-1,1]$.

\paragraph{Max Sensitivity (Robustness)}
Max Sensitivity evaluates how much an explanation changes under small input perturbations~\citep{yeh2019infidelitysensitivityexplanations}.
Given perturbations $\delta_k\sim\mathcal{D}_\epsilon$ supported on a small $\ell_p$-ball, e.g., $\|\delta_k\|_\infty\le\epsilon$, it is defined as
\begin{equation}
\mathrm{MaxSens}(\vx)
=
\max_{k\in\{1,\dots,K\}}
\frac{\|e(\vx)-e(\vx+\delta_k)\|}{\|e(\vx)\|}
\end{equation}
Lower Max Sensitivity indicates more robust explanations.

We use this metric to test whether explanations remain stable under small, visually minor perturbations of the input.
In our Quantus configuration, we set the perturbation \texttt{lower\_bound} to $0.02$, matching the scale of the default perturbation function more closely and yielding a meaningful small-perturbation robustness test.

\paragraph{Random Logit (Randomisation)}
Random Logit compares the explanation for the true class with the explanation for a randomly chosen non-target class~\citep{sixt2024explanationsliemodifiedbp}.
For each sample, choose $y_{\mathrm{rand}}\in\{1,\dots,C\}\setminus\{y\}$ uniformly at random and compute a reference explanation $e(\vx;y_{\mathrm{rand}})$.
We define
\begin{equation}
\mathrm{RandLogit}(\vx)
=
|\rho_{\mathrm{Pearson}}
\big(
e(\vx;y),
e(\vx;y_{\mathrm{rand}})
\big)|
\end{equation}
Lower Random Logit is better, since explanations for different classes should be meaningfully different.

Quantus uses SSIM by default for this metric.
We replace it with Pearson correlation because it directly measures whether two explanation vectors point in similar input-space directions.
We use Random Logit as a sanity check for target specificity rather than as a faithfulness metric.

\section{Hyperparameter Robustness}
\label{app:hyperparameter_sensitivity}

We assessed hyperparameter robustness through two complementary analyses: exploratory quantitative and qualitative experiments on smaller validation subsets during method development, and controlled one-factor-at-a-time ablations on 1000 ImageNet validation images, with 1000 ImageNet-derived mosaics for Focus.

\paragraph{Exploratory analyses and feature accentuation}
For ReLU-based architectures, we explored alternative surrogate gating functions, including sigmoid and Normal cumulative distribution function (CDF) gates.
These development-time experiments suggested qualitatively similar behaviour across gate families and moderate temperature ranges, although more aggressive softening could degrade explanations.
For MaxPool, we explored temperatures between $0.01$ and $0.5$.
For PVT, meaningful visual explanations were obtained across a broad range of attention temperatures, including values as large as $15$.

For PullbackAscent, the step size $\alpha$ and number of steps $K$ also control the appearance of feature accentuations.
Fewer or smaller steps keep the modified image closer to the source, whereas moderate ascent tends to make target-specific structures more pronounced and visually coherent.
We use $\alpha=20$ and $K=5$, with $\ell_2$-normalised ascent directions, as an empirical compromise between locality, perceptual clarity, and quantitative performance.

\paragraph{Controlled ablations}
Within each architecture, ablation configurations reuse the same evaluation examples, and random seeds are reset between configurations.
Each ablation changes one parameter while retaining
\begin{equation}
\begin{gathered}
\tau_{\mathrm{ReLU}}=0.6,\qquad
\tau_{\mathrm{MaxPool}}=0.3,\qquad
\tau_{\mathrm{GELU}}=\tau_{\mathrm{Attn}}=1,\\
\alpha=20,\qquad K=5.
\end{gathered}
\end{equation}
Here, the temperature subscripts identify the corresponding operators.
Temperature ablations apply to both SoftPullback (SoftPb) and
PullbackAscent (PA), whereas $K$ and $\alpha$ apply only to PA.
Only temperatures relevant to each architecture are varied.

Tables~\ref{table:ablations_resnet50}, \ref{table:ablations_pvt},
and~\ref{table:ablations_vgg} report ablation means and standard
deviations, with the default-setting results from the main evaluation.

\paragraph{Surrogate temperatures}
The temperature sweeps reveal different trade-offs on either side
of the defaults.
For the CNNs, reducing $\tau_{\mathrm{ReLU}}$ below $0.6$
generally increases Infidelity and Max Sensitivity, with the largest
Infidelity increases occurring for SoftPb at $0.3$.
Focus also tends to decrease, although intermediate temperatures
can improve correlation-based faithfulness.
Increasing $\tau_{\mathrm{ReLU}}$ towards $1$ instead reduces
Max Sensitivity, but lowers both correlation scores on both
architectures; its effects on Infidelity and Focus are mixed.

Reducing $\tau_{\mathrm{MaxPool}}$ towards $0.01$ improves Infidelity but lowers Focus and worsens Max Sensitivity on both CNNs, particularly on VGG.
Increasing it towards $0.5$ instead improves Max Sensitivity while increasing Infidelity.
These results reveal a trade-off between Infidelity and local stability, with the defaults providing intermediate settings.

For PVT, both methods show relatively stable performance under moderate variations in GELU and attention temperatures, although some metrics exhibit more pronounced changes.
Increasing $\tau_{\mathrm{GELU}}$ to $2$ consistently worsens all five primary metrics for both methods.
Increasing $\tau_{\mathrm{Attn}}$ to $5$ improves Infidelity and Max Sensitivity but reduces Focus and both correlation scores.
Overall, the results indicate a degree of robustness to temperature variations, alongside metric-dependent trade-offs.

\paragraph{Ascent parameters}
Reducing $K$ below the default of $5$ progressively increases Infidelity on PVT and VGG, whereas changes on ResNet50 are comparatively small.
On the CNNs, fewer steps improve Max Sensitivity, while $K=2$ or $3$ also improves Focus.
On PVT, fewer steps increase Focus but worsen Max Sensitivity.
Increasing $K$ to $10$ slightly improves Infidelity on PVT and VGG, but reduces Focus and Faithfulness Correlation on all three architectures.
Max Sensitivity improves on PVT but worsens on CNNs.

Together with the qualitative observations, these sweeps show
that moving towards more pronounced feature accentuations
can trade localisation or stability for improvements in
other metrics.

\begin{table}[tb]
\small
\setlength{\tabcolsep}{3pt}
\caption{ResNet50 hyperparameter ablations on 1000 ImageNet validation samples and 1000.
Default settings are $\tau_{\mathrm{ReLU}}=0.6$,
$\tau_{\mathrm{MaxPool}}=0.3$, $K=5$, and $\alpha=20$;
$K$ and $\alpha$ apply only to PA.
Each ablation changes only the indicated parameter.
Entries report means and standard deviations.
}
\label{table:ablations_resnet50}
\begin{center}
\begin{tabular}{llrrrrrr}
\toprule
Method & Setting
& Infidelity $\downarrow$
& Focus $\uparrow$
& Faith.Corr $\uparrow$
& Mono.Corr $\uparrow$
& Max.Sens $\downarrow$
& Rand.Logit $\downarrow$ \\
\midrule
SoftPb & \textbf{Default}
& $5.989_{\pm 5.67}$ & $0.746_{\pm 0.13}$ & $0.389_{\pm 0.37}$ & $0.420_{\pm 0.39}$ & $0.119_{\pm 0.04}$ & $0.062_{\pm 0.42}$ \\
SoftPb & $\tau_{\mathrm{ReLU}}=0.3$
& $25.518_{\pm 33.9}$ & $0.687_{\pm 0.12}$ & $0.383_{\pm 0.37}$ & $0.412_{\pm 0.37}$ & $0.197_{\pm 0.06}$ & $0.010_{\pm 0.26}$ \\
SoftPb & $\tau_{\mathrm{ReLU}}=0.4$
& $7.814_{\pm 10.0}$ & $0.721_{\pm 0.12}$ & $0.419_{\pm 0.35}$ & $0.424_{\pm 0.38}$ & $0.158_{\pm 0.05}$ & $0.002_{\pm 0.32}$ \\
SoftPb & $\tau_{\mathrm{ReLU}}=0.5$
& $6.095_{\pm 5.98}$ & $0.740_{\pm 0.13}$ & $0.410_{\pm 0.36}$ & $0.425_{\pm 0.38}$ & $0.132_{\pm 0.04}$ & $0.003_{\pm 0.39}$ \\
SoftPb & $\tau_{\mathrm{ReLU}}=1$
& $6.551_{\pm 6.64}$ & $0.717_{\pm 0.14}$ & $0.331_{\pm 0.40}$ & $0.375_{\pm 0.40}$ & $0.105_{\pm 0.03}$ & $0.012_{\pm 0.45}$ \\
SoftPb & $\tau_{\mathrm{MaxPool}}=0.01$
& $5.842_{\pm 5.71}$ & $0.714_{\pm 0.13}$ & $0.388_{\pm 0.38}$ & $0.419_{\pm 0.38}$ & $0.332_{\pm 0.08}$ & $0.006_{\pm 0.42}$ \\
SoftPb & $\tau_{\mathrm{MaxPool}}=0.5$
& $6.082_{\pm 5.89}$ & $0.748_{\pm 0.13}$ & $0.392_{\pm 0.37}$ & $0.421_{\pm 0.38}$ & $0.114_{\pm 0.04}$ & $0.007_{\pm 0.43}$ \\
\midrule
PA & \textbf{Default}
& $5.384_{\pm 4.83}$ & $0.726_{\pm 0.13}$ & $0.382_{\pm 0.37}$ & $0.421_{\pm 0.40}$ & $0.244_{\pm 0.09}$ & $0.212_{\pm 0.19}$ \\
PA & $\tau_{\mathrm{ReLU}}=0.3$
& $8.350_{\pm 10.4}$ & $0.639_{\pm 0.11}$ & $0.380_{\pm 0.35}$ & $0.364_{\pm 0.37}$ & $0.563_{\pm 0.13}$ & $0.067_{\pm 0.09}$ \\
PA & $\tau_{\mathrm{ReLU}}=0.4$
& $5.650_{\pm 4.85}$ & $0.665_{\pm 0.11}$ & $0.403_{\pm 0.33}$ & $0.392_{\pm 0.38}$ & $0.364_{\pm 0.11}$ & $0.144_{\pm 0.13}$ \\
PA & $\tau_{\mathrm{ReLU}}=0.5$
& $5.354_{\pm 4.73}$ & $0.700_{\pm 0.12}$ & $0.398_{\pm 0.35}$ & $0.423_{\pm 0.39}$ & $0.271_{\pm 0.09}$ & $0.203_{\pm 0.17}$ \\
PA & $\tau_{\mathrm{ReLU}}=1$
& $5.727_{\pm 5.31}$ & $0.733_{\pm 0.14}$ & $0.332_{\pm 0.40}$ & $0.401_{\pm 0.40}$ & $0.233_{\pm 0.10}$ & $0.234_{\pm 0.24}$ \\
PA & $\tau_{\mathrm{MaxPool}}=0.01$
& $5.338_{\pm 4.86}$ & $0.675_{\pm 0.12}$ & $0.368_{\pm 0.37}$ & $0.361_{\pm 0.39}$ & $0.489_{\pm 0.13}$ & $0.200_{\pm 0.19}$ \\
PA & $\tau_{\mathrm{MaxPool}}=0.5$
& $5.405_{\pm 4.86}$ & $0.733_{\pm 0.13}$ & $0.387_{\pm 0.37}$ & $0.431_{\pm 0.40}$ & $0.217_{\pm 0.09}$ & $0.230_{\pm 0.20}$ \\
\addlinespace[3pt]
PA & $K=1$
& $5.955_{\pm 5.63}$ & $0.746_{\pm 0.13}$ & $0.389_{\pm 0.37}$ & $0.418_{\pm 0.39}$ & $0.115_{\pm 0.03}$ & $0.006_{\pm 0.43}$ \\
PA & $K=2$
& $5.471_{\pm 5.01}$ & $0.757_{\pm 0.13}$ & $0.396_{\pm 0.37}$ & $0.411_{\pm 0.39}$ & $0.139_{\pm 0.05}$ & $0.152_{\pm 0.31}$ \\
PA & $K=3$
& $5.352_{\pm 4.86}$ & $0.750_{\pm 0.13}$ & $0.393_{\pm 0.37}$ & $0.418_{\pm 0.40}$ & $0.179_{\pm 0.07}$ & $0.199_{\pm 0.26}$ \\
PA & $K=10$
& $5.527_{\pm 5.05}$ & $0.674_{\pm 0.13}$ & $0.358_{\pm 0.36}$ & $0.423_{\pm 0.40}$ & $0.358_{\pm 0.11}$ & $0.219_{\pm 0.14}$ \\
\addlinespace[3pt]
PA & $\alpha=5$
& $5.393_{\pm 4.94}$ & $0.773_{\pm 0.13}$ & $0.403_{\pm 0.37}$ & $0.418_{\pm 0.40}$ & $0.179_{\pm 0.08}$ & $0.261_{\pm 0.25}$ \\
PA & $\alpha=10$
& $5.327_{\pm 4.85}$ & $0.756_{\pm 0.13}$ & $0.396_{\pm 0.37}$ & $0.420_{\pm 0.40}$ & $0.208_{\pm 0.09}$ & $0.261_{\pm 0.22}$ \\
PA & $\alpha=40$
& $5.472_{\pm 5.08}$ & $0.688_{\pm 0.13}$ & $0.362_{\pm 0.37}$ & $0.415_{\pm 0.40}$ & $0.305_{\pm 0.10}$ & $0.173_{\pm 0.19}$ \\
\bottomrule
\end{tabular}
\end{center}
\end{table}

\begin{table}[tb]
\small
\setlength{\tabcolsep}{3pt}
\caption{PVT hyperparameter ablations on 1000 ImageNet validation samples.
Default settings are $\tau_{\mathrm{GELU}}=\tau_{\mathrm{Attn}}=1$,
$K=5$, and $\alpha=20$; $K$ and $\alpha$ apply only to PA.}
\label{table:ablations_pvt}
\begin{center}
\begin{tabular}{llrrrrrr}
\toprule
Method & Setting
& Infidelity $\downarrow$
& Focus $\uparrow$
& Faith.Corr $\uparrow$
& Mono.Corr $\uparrow$
& Max.Sens $\downarrow$
& Rand.Logit $\downarrow$ \\
\midrule
SoftPb & \textbf{Default}
& $6.264_{\pm 6.40}$ & $0.617_{\pm 0.18}$ & $0.119_{\pm 0.34}$ & $0.164_{\pm 0.43}$ & $1.066_{\pm 0.18}$ & $0.006_{\pm 0.35}$ \\
SoftPb & $\tau_{\mathrm{GELU}}=0.5$
& $6.670_{\pm 6.23}$ & $0.637_{\pm 0.17}$ & $0.142_{\pm 0.34}$ & $0.199_{\pm 0.43}$ & $0.998_{\pm 0.14}$ & $0.008_{\pm 0.32}$ \\
SoftPb & $\tau_{\mathrm{GELU}}=2$
& $7.099_{\pm 7.04}$ & $0.548_{\pm 0.21}$ & $0.066_{\pm 0.30}$ & $0.017_{\pm 0.45}$ & $1.216_{\pm 0.30}$ & $0.041_{\pm 0.51}$ \\
SoftPb & $\tau_{\mathrm{Attn}}=0.5$
& $7.115_{\pm 6.58}$ & $0.616_{\pm 0.18}$ & $0.104_{\pm 0.33}$ & $0.170_{\pm 0.43}$ & $1.094_{\pm 0.17}$ & $0.024_{\pm 0.37}$ \\
SoftPb & $\tau_{\mathrm{Attn}}=5$
& $5.944_{\pm 6.60}$ & $0.572_{\pm 0.17}$ & $0.107_{\pm 0.31}$ & $0.079_{\pm 0.44}$ & $1.039_{\pm 0.15}$ & $0.020_{\pm 0.36}$ \\
\midrule
PA & \textbf{Default}
& $1.634_{\pm 1.03}$ & $0.554_{\pm 0.10}$ & $0.122_{\pm 0.36}$ & $0.219_{\pm 0.33}$ & $0.855_{\pm 0.12}$ & $0.121_{\pm 0.11}$ \\
PA & $\tau_{\mathrm{GELU}}=0.5$
& $1.592_{\pm 0.99}$ & $0.557_{\pm 0.10}$ & $0.129_{\pm 0.36}$ & $0.222_{\pm 0.33}$ & $0.890_{\pm 0.12}$ & $0.108_{\pm 0.11}$ \\
PA & $\tau_{\mathrm{GELU}}=2$
& $3.134_{\pm 3.92}$ & $0.526_{\pm 0.10}$ & $0.090_{\pm 0.36}$ & $0.185_{\pm 0.33}$ & $0.888_{\pm 0.12}$ & $0.060_{\pm 0.14}$ \\
PA & $\tau_{\mathrm{Attn}}=0.5$
& $1.681_{\pm 1.10}$ & $0.561_{\pm 0.10}$ & $0.120_{\pm 0.36}$ & $0.220_{\pm 0.35}$ & $0.927_{\pm 0.11}$ & $0.101_{\pm 0.10}$ \\
PA & $\tau_{\mathrm{Attn}}=5$
& $1.569_{\pm 1.02}$ & $0.538_{\pm 0.09}$ & $0.109_{\pm 0.35}$ & $0.209_{\pm 0.31}$ & $0.854_{\pm 0.14}$ & $0.112_{\pm 0.12}$ \\
\addlinespace[3pt]
PA & $K=1$
& $6.224_{\pm 6.35}$ & $0.617_{\pm 0.18}$ & $0.119_{\pm 0.34}$ & $0.164_{\pm 0.43}$ & $1.110_{\pm 0.16}$ & $0.006_{\pm 0.35}$ \\
PA & $K=2$
& $2.686_{\pm 2.78}$ & $0.591_{\pm 0.12}$ & $0.134_{\pm 0.35}$ & $0.215_{\pm 0.38}$ & $1.008_{\pm 0.13}$ & $0.044_{\pm 0.20}$ \\
PA & $K=3$
& $1.926_{\pm 1.64}$ & $0.574_{\pm 0.11}$ & $0.134_{\pm 0.36}$ & $0.224_{\pm 0.35}$ & $0.939_{\pm 0.12}$ & $0.073_{\pm 0.15}$ \\
PA & $K=10$
& $1.556_{\pm 0.96}$ & $0.537_{\pm 0.09}$ & $0.103_{\pm 0.36}$ & $0.221_{\pm 0.33}$ & $0.800_{\pm 0.12}$ & $0.139_{\pm 0.07}$ \\
\addlinespace[3pt]
PA & $\alpha=5$
& $1.962_{\pm 1.66}$ & $0.571_{\pm 0.12}$ & $0.133_{\pm 0.36}$ & $0.192_{\pm 0.38}$ & $0.850_{\pm 0.13}$ & $0.111_{\pm 0.13}$ \\
PA & $\alpha=10$
& $1.702_{\pm 1.11}$ & $0.560_{\pm 0.11}$ & $0.130_{\pm 0.36}$ & $0.209_{\pm 0.35}$ & $0.849_{\pm 0.13}$ & $0.114_{\pm 0.11}$ \\
PA & $\alpha=40$
& $1.616_{\pm 1.04}$ & $0.553_{\pm 0.09}$ & $0.110_{\pm 0.36}$ & $0.227_{\pm 0.33}$ & $0.867_{\pm 0.11}$ & $0.105_{\pm 0.10}$ \\
\bottomrule
\end{tabular}
\end{center}
\end{table}

\begin{table}[tb]
\small
\setlength{\tabcolsep}{3pt}
\caption{VGG hyperparameter ablations on 1000 ImageNet validation
samples.
Default settings are $\tau_{\mathrm{ReLU}}=0.6$,
$\tau_{\mathrm{MaxPool}}=0.3$, $K=5$, and $\alpha=20$;
$K$ and $\alpha$ apply only to PA.}
\label{table:ablations_vgg}
\begin{center}
\begin{tabular}{llrrrrrr}
\toprule
Method & Setting
& Infidelity $\downarrow$
& Focus $\uparrow$
& Faith.Corr $\uparrow$
& Mono.Corr $\uparrow$
& Max.Sens $\downarrow$
& Rand.Logit $\downarrow$ \\
\midrule
SoftPb & \textbf{Default}
& $16.303_{\pm 31.68}$ & $0.614_{\pm 0.11}$ & $0.606_{\pm 0.30}$ & $0.375_{\pm 0.39}$ & $0.212_{\pm 0.08}$ & $0.064_{\pm 0.33}$ \\
SoftPb & $\tau_{\mathrm{ReLU}}=0.3$
& $38.167_{\pm 56.98}$ & $0.588_{\pm 0.12}$ & $0.578_{\pm 0.32}$ & $0.401_{\pm 0.38}$ & $0.245_{\pm 0.07}$ & $0.008_{\pm 0.31}$ \\
SoftPb & $\tau_{\mathrm{ReLU}}=0.4$
& $25.580_{\pm 40.69}$ & $0.600_{\pm 0.12}$ & $0.593_{\pm 0.31}$ & $0.391_{\pm 0.38}$ & $0.229_{\pm 0.07}$ & $0.008_{\pm 0.31}$ \\
SoftPb & $\tau_{\mathrm{ReLU}}=0.5$
& $19.038_{\pm 33.67}$ & $0.610_{\pm 0.11}$ & $0.603_{\pm 0.30}$ & $0.382_{\pm 0.39}$ & $0.219_{\pm 0.07}$ & $0.007_{\pm 0.31}$ \\
SoftPb & $\tau_{\mathrm{ReLU}}=1$
& $14.571_{\pm 32.76}$ & $0.609_{\pm 0.11}$ & $0.577_{\pm 0.31}$ & $0.323_{\pm 0.41}$ & $0.203_{\pm 0.09}$ & $0.002_{\pm 0.35}$ \\
SoftPb & $\tau_{\mathrm{MaxPool}}=0.01$
& $9.141_{\pm 24.72}$ & $0.596_{\pm 0.11}$ & $0.596_{\pm 0.29}$ & $0.399_{\pm 0.39}$ & $0.555_{\pm 0.15}$ & $0.003_{\pm 0.31}$ \\
SoftPb & $\tau_{\mathrm{MaxPool}}=0.5$
& $24.132_{\pm 43.56}$ & $0.611_{\pm 0.11}$ & $0.603_{\pm 0.30}$ & $0.357_{\pm 0.39}$ & $0.208_{\pm 0.08}$ & $0.004_{\pm 0.31}$ \\
\midrule
PA & \textbf{Default}
& $5.616_{\pm 13.18}$ & $0.623_{\pm 0.11}$ & $0.570_{\pm 0.29}$ & $0.312_{\pm 0.37}$ & $0.390_{\pm 0.11}$ & $0.136_{\pm 0.13}$ \\
PA & $\tau_{\mathrm{ReLU}}=0.3$
& $6.452_{\pm 13.46}$ & $0.619_{\pm 0.10}$ & $0.571_{\pm 0.29}$ & $0.314_{\pm 0.36}$ & $0.557_{\pm 0.12}$ & $0.089_{\pm 0.12}$ \\
PA & $\tau_{\mathrm{ReLU}}=0.4$
& $5.700_{\pm 10.47}$ & $0.623_{\pm 0.10}$ & $0.572_{\pm 0.29}$ & $0.321_{\pm 0.36}$ & $0.482_{\pm 0.12}$ & $0.117_{\pm 0.12}$ \\
PA & $\tau_{\mathrm{ReLU}}=0.5$
& $5.636_{\pm 12.15}$ & $0.625_{\pm 0.11}$ & $0.572_{\pm 0.29}$ & $0.330_{\pm 0.36}$ & $0.427_{\pm 0.11}$ & $0.144_{\pm 0.13}$ \\
PA & $\tau_{\mathrm{ReLU}}=1$
& $6.313_{\pm 16.17}$ & $0.605_{\pm 0.10}$ & $0.519_{\pm 0.31}$ & $0.285_{\pm 0.40}$ & $0.329_{\pm 0.13}$ & $0.216_{\pm 0.17}$ \\
PA & $\tau_{\mathrm{MaxPool}}=0.01$
& $4.097_{\pm 04.51}$ & $0.577_{\pm 0.09}$ & $0.530_{\pm 0.29}$ & $0.289_{\pm 0.33}$ & $0.765_{\pm 0.13}$ & $0.133_{\pm 0.13}$ \\
PA & $\tau_{\mathrm{MaxPool}}=0.5$
& $7.305_{\pm 18.13}$ & $0.627_{\pm 0.11}$ & $0.572_{\pm 0.30}$ & $0.314_{\pm 0.36}$ & $0.346_{\pm 0.11}$ & $0.168_{\pm 0.14}$ \\
\addlinespace[3pt]
PA & $K=1$
& $15.933_{\pm 29.73}$ & $0.615_{\pm 0.11}$ & $0.606_{\pm 0.30}$ & $0.366_{\pm 0.39}$ & $0.208_{\pm 0.07}$ & $0.005_{\pm 0.32}$ \\
PA & $K=2$
& $8.998_{\pm 20.86}$ & $0.637_{\pm 0.11}$ & $0.601_{\pm 0.29}$ & $0.329_{\pm 0.38}$ & $0.251_{\pm 0.08}$ & $0.101_{\pm 0.21}$ \\
PA & $K=3$
& $6.798_{\pm 17.18}$ & $0.638_{\pm 0.11}$ & $0.594_{\pm 0.29}$ & $0.320_{\pm 0.38}$ & $0.304_{\pm 0.10}$ & $0.136_{\pm 0.18}$ \\
PA & $K=10$
& $5.090_{\pm 08.63}$ & $0.606_{\pm 0.10}$ & $0.537_{\pm 0.30}$ & $0.315_{\pm 0.36}$ & $0.521_{\pm 0.13}$ & $0.165_{\pm 0.10}$ \\
\addlinespace[3pt]
PA & $\alpha=5$
& $6.321_{\pm 13.88}$ & $0.647_{\pm 0.11}$ & $0.588_{\pm 0.29}$ & $0.321_{\pm 0.39}$ & $0.249_{\pm 0.09}$ & $0.211_{\pm 0.17}$ \\
PA & $\alpha=10$
& $5.738_{\pm 13.28}$ & $0.638_{\pm 0.11}$ & $0.577_{\pm 0.29}$ & $0.314_{\pm 0.38}$ & $0.299_{\pm 0.09}$ & $0.206_{\pm 0.15}$ \\
PA & $\alpha=40$
& $6.150_{\pm 15.74}$ & $0.611_{\pm 0.10}$ & $0.570_{\pm 0.29}$ & $0.319_{\pm 0.36}$ & $0.538_{\pm 0.12}$ & $0.112_{\pm 0.13}$ \\
\bottomrule
\end{tabular}
\end{center}
\end{table}
\section{Additional figures and tables}\label{app:figures_and_tables}

\begin{figure}[tbh]
\begin{center}
\includegraphics[width=0.75\linewidth]
{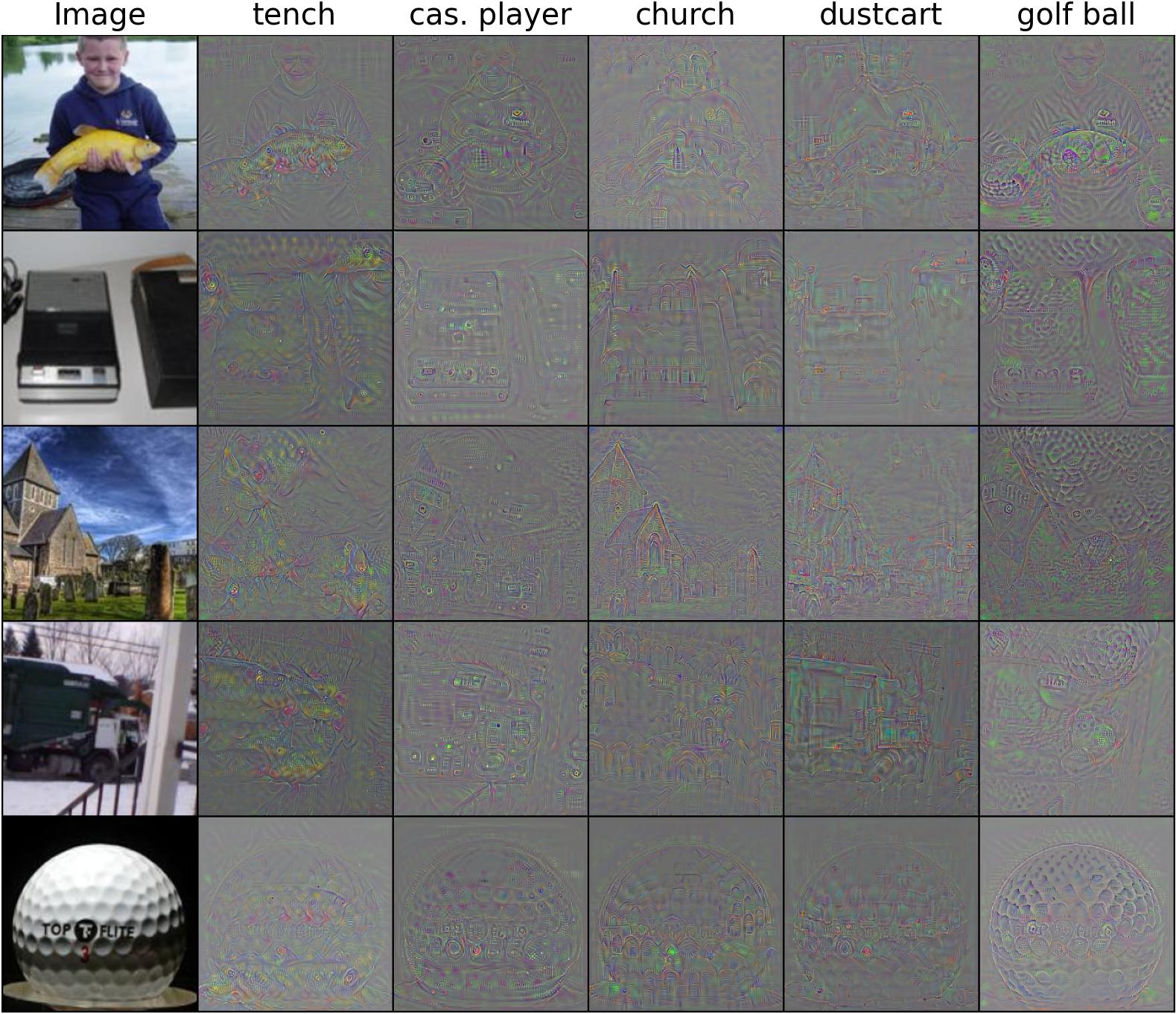}


\includegraphics[width=0.75\linewidth]
{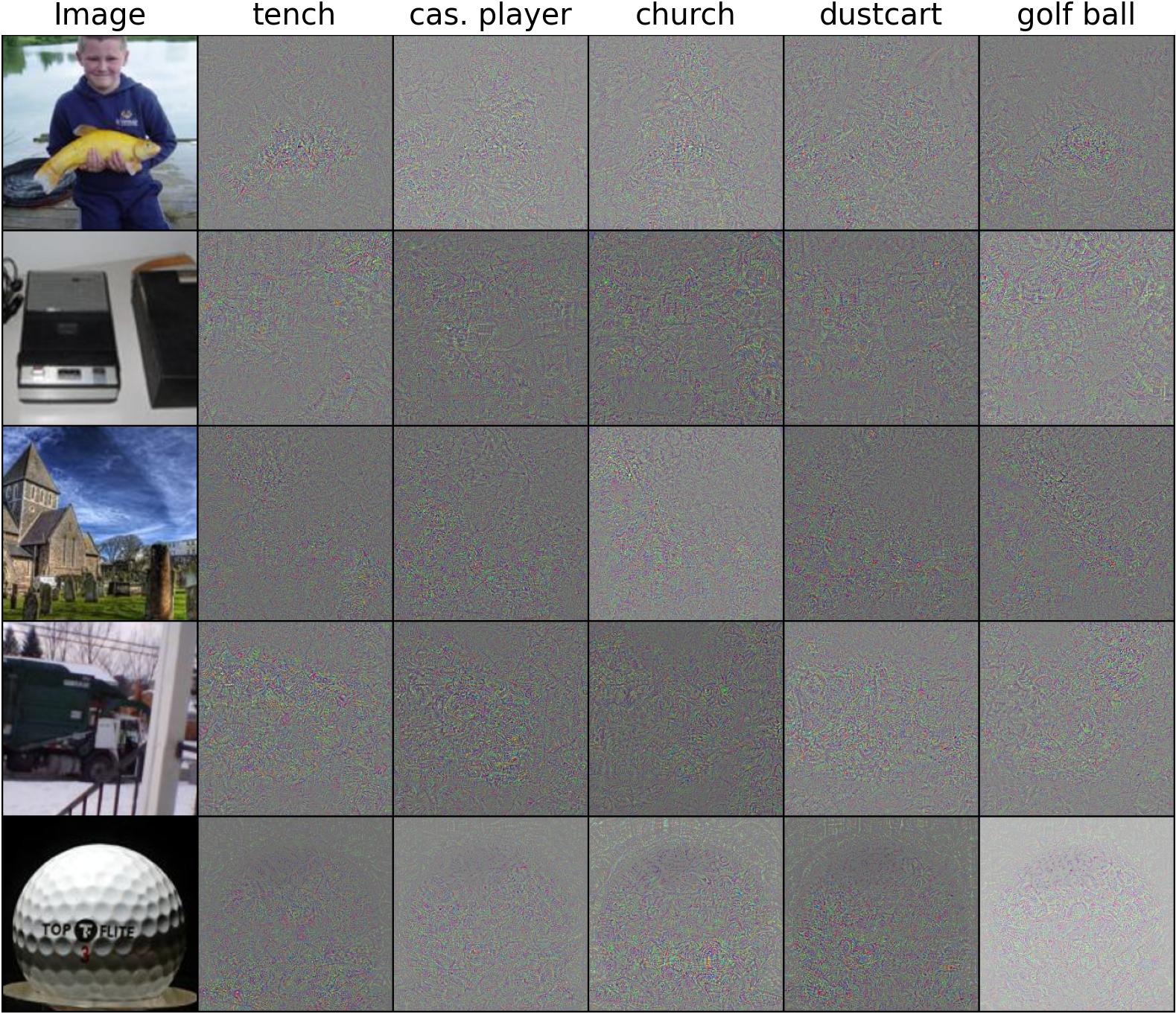}

\caption{Class-specific accentuations for VGG using local ascent with default parameters.
\textbf{Top}: by PullbackAscent.
\textbf{Bottom}: by GradientAscent.}
\label{fig:counterfactuals_vgg}
\end{center}
\end{figure}

\begin{figure}[tbh]
\begin{center}
\includegraphics[width=0.75\linewidth]
{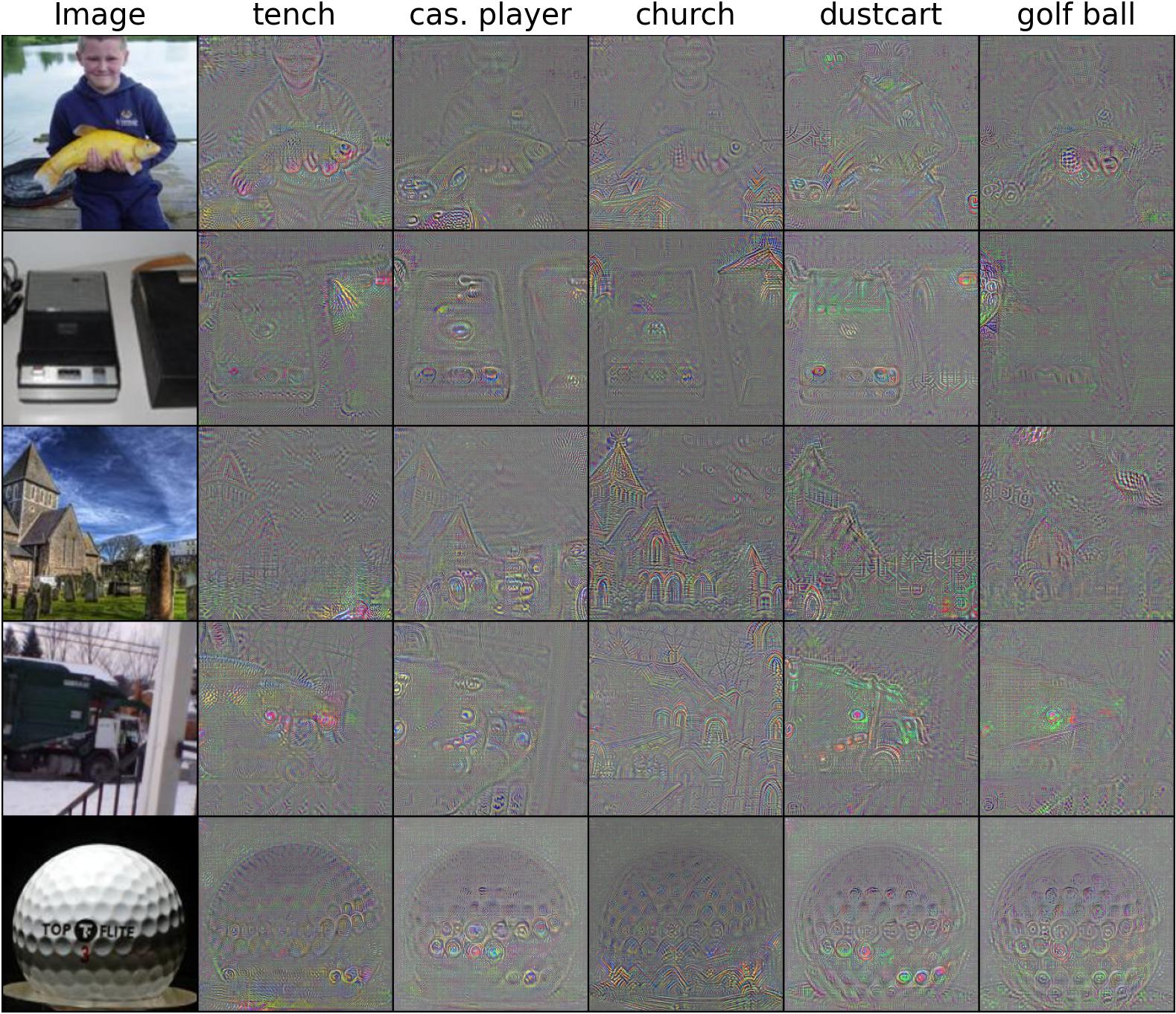}


\includegraphics[width=0.75\linewidth]
{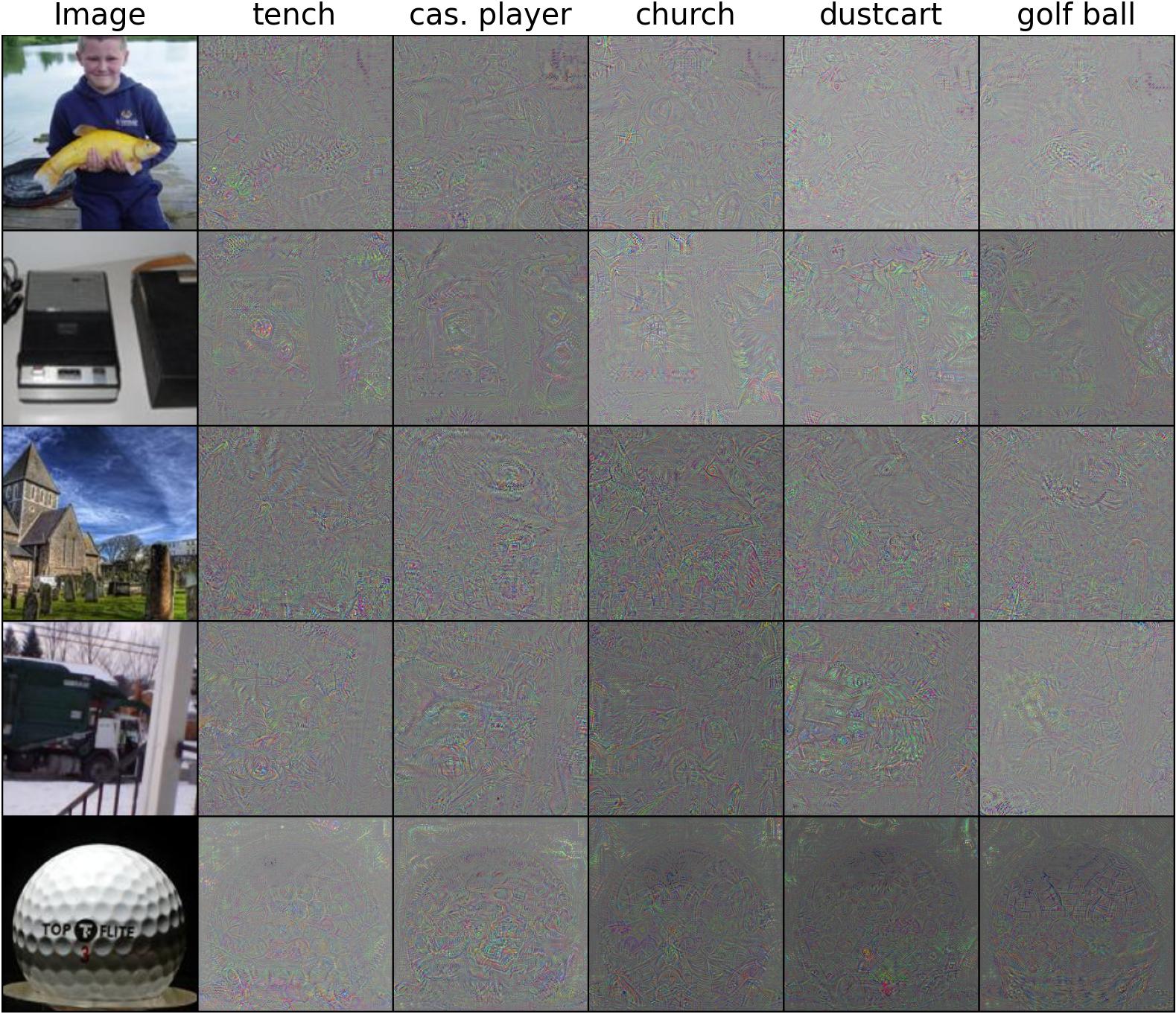}

\caption{Class-specific accentuations for PVT using local ascent with $K=10$ steps to emphasize visual structure.
\textbf{Top}: by PullbackAscent.
\textbf{Bottom}: by GradientAscent.}
\label{fig:counterfactuals_pvt}
\end{center}
\end{figure}

\begin{figure}[tb]
  \begin{center}
    \centerline{\includegraphics[width=0.83\linewidth]{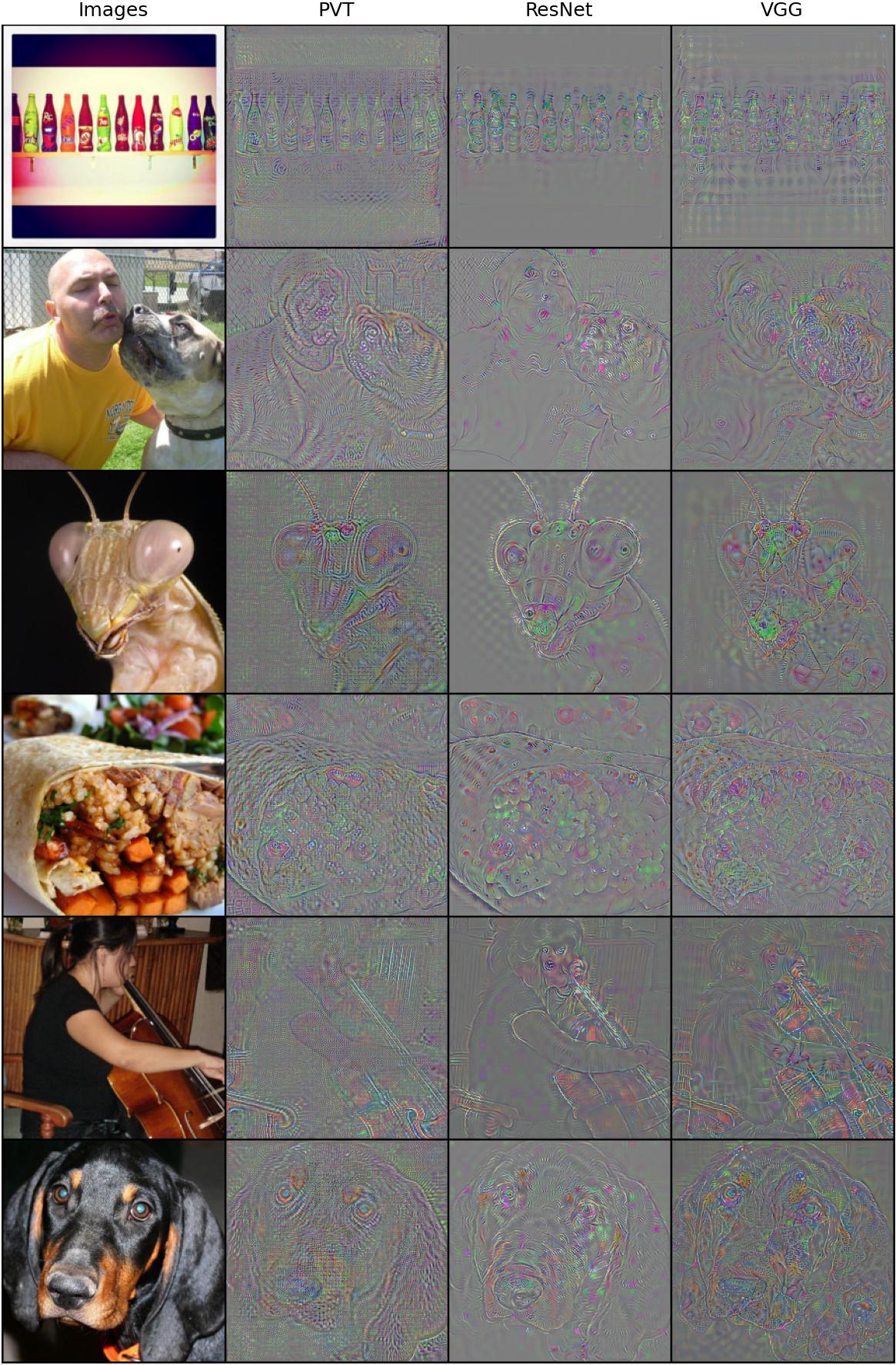}}
    \caption{Target class accentuations (true label, cf. Section~\ref{sec:accentuations}) for all studied models using 10-step PullbackAscent (cf. Section~\ref{sec:pullback_ascent}).}
    \label{fig:accentuations_true_label}
  \end{center}
\end{figure}

\begin{figure}[tb]
  \begin{center}
    \centerline{\includegraphics[width=0.83\linewidth]{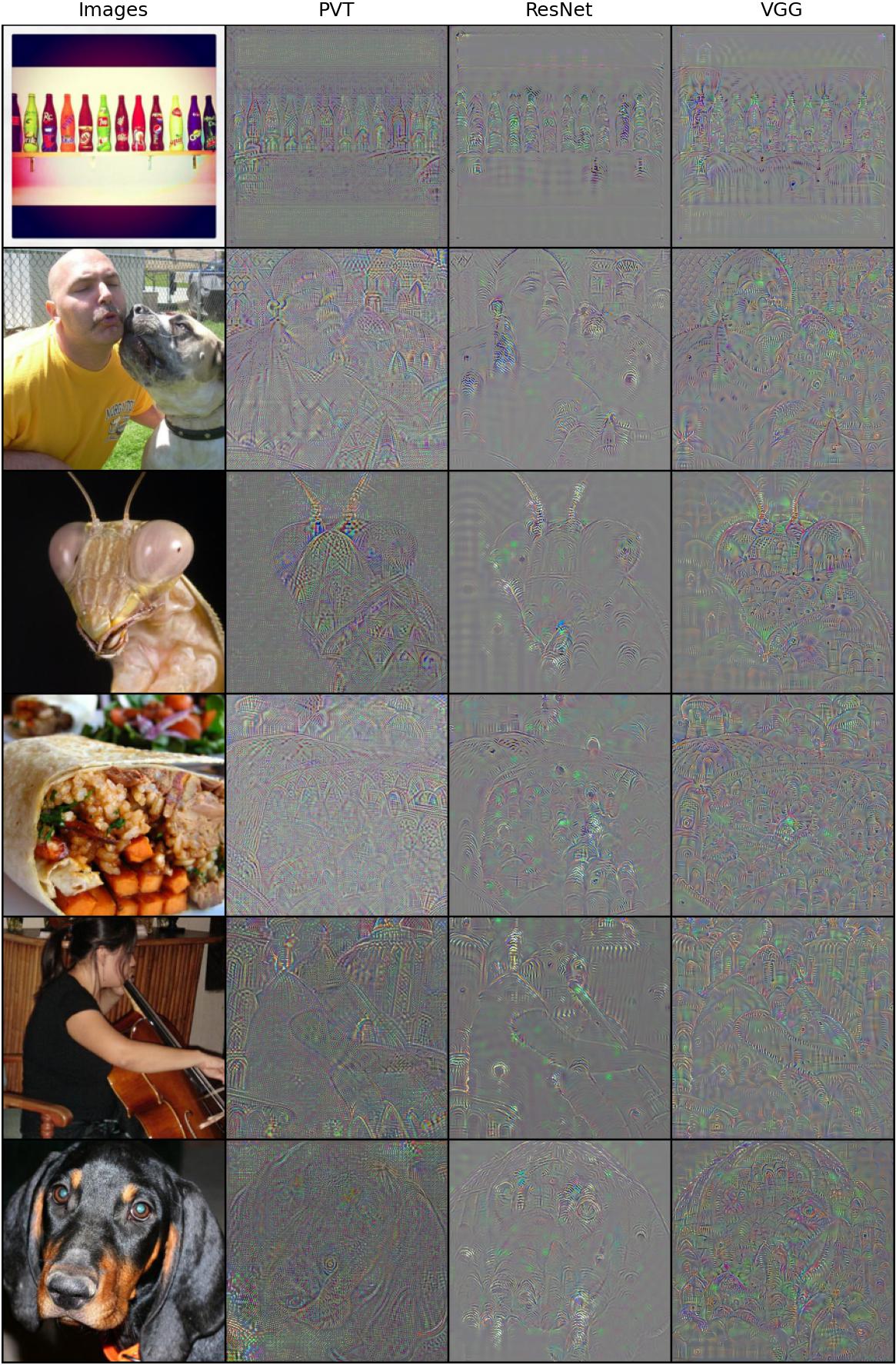}}
    \caption{Class accentuations (cf. Section~\ref{sec:accentuations}) for target class ``mosque'' for all studied models using 10-step PullbackAscent (cf. Section~\ref{sec:pullback_ascent}).}
    \label{fig:accentuations_mosque}
  \end{center}
\end{figure}

\begin{figure}[tb]
  \begin{center}
    \centerline{\includegraphics[width=0.83\linewidth]{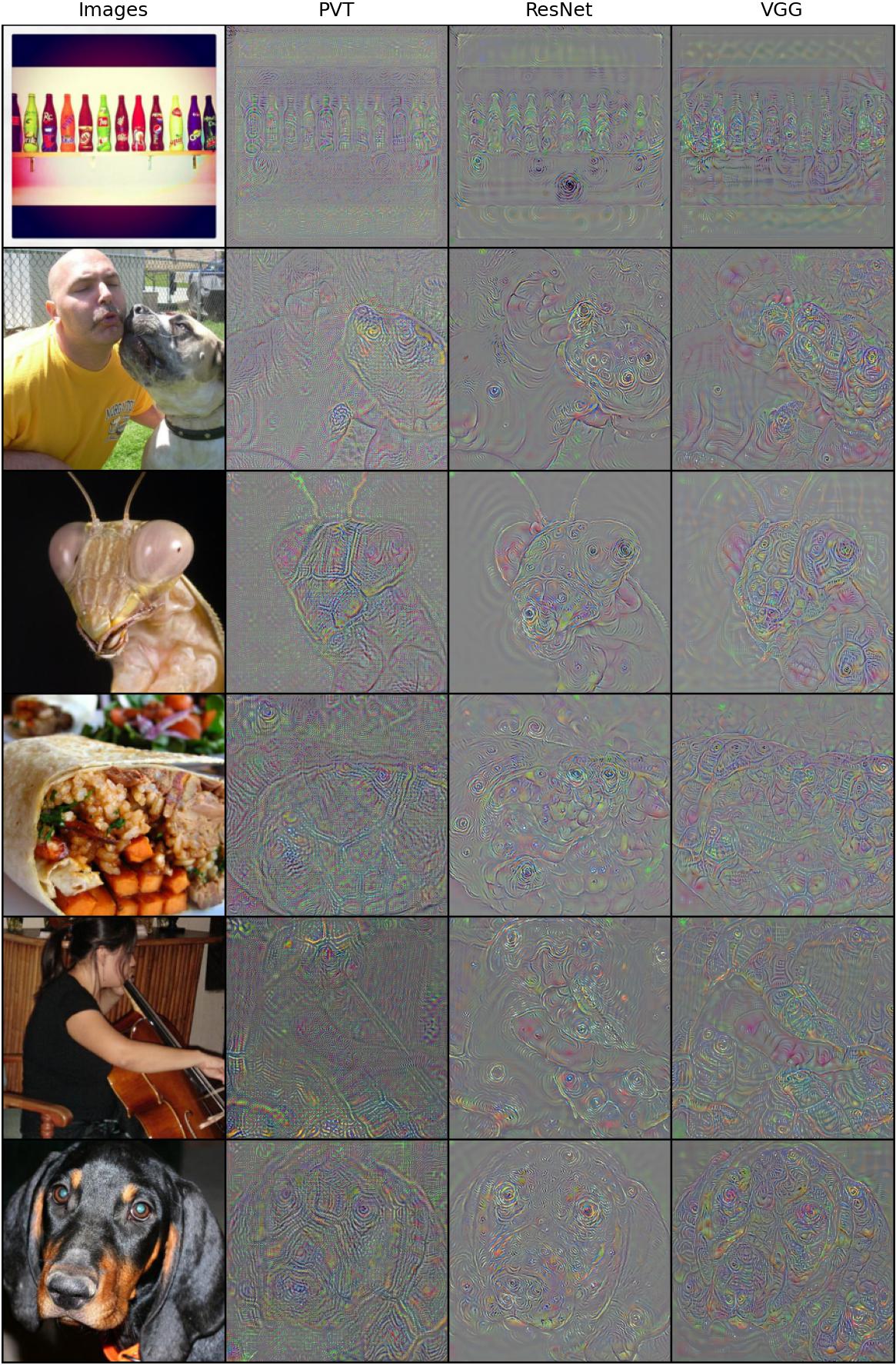}}
    \caption{Class accentuations (cf. Section~\ref{sec:accentuations}) for target class ``mud turtle'' for all studied models using 10-step PullbackAscent (cf. Section~\ref{sec:pullback_ascent}).}
    \label{fig:accentuations_mud_turtle}
  \end{center}
\end{figure}

\begin{figure}[tb]
  \begin{center}
    \centerline{\includegraphics[width=0.83\linewidth]{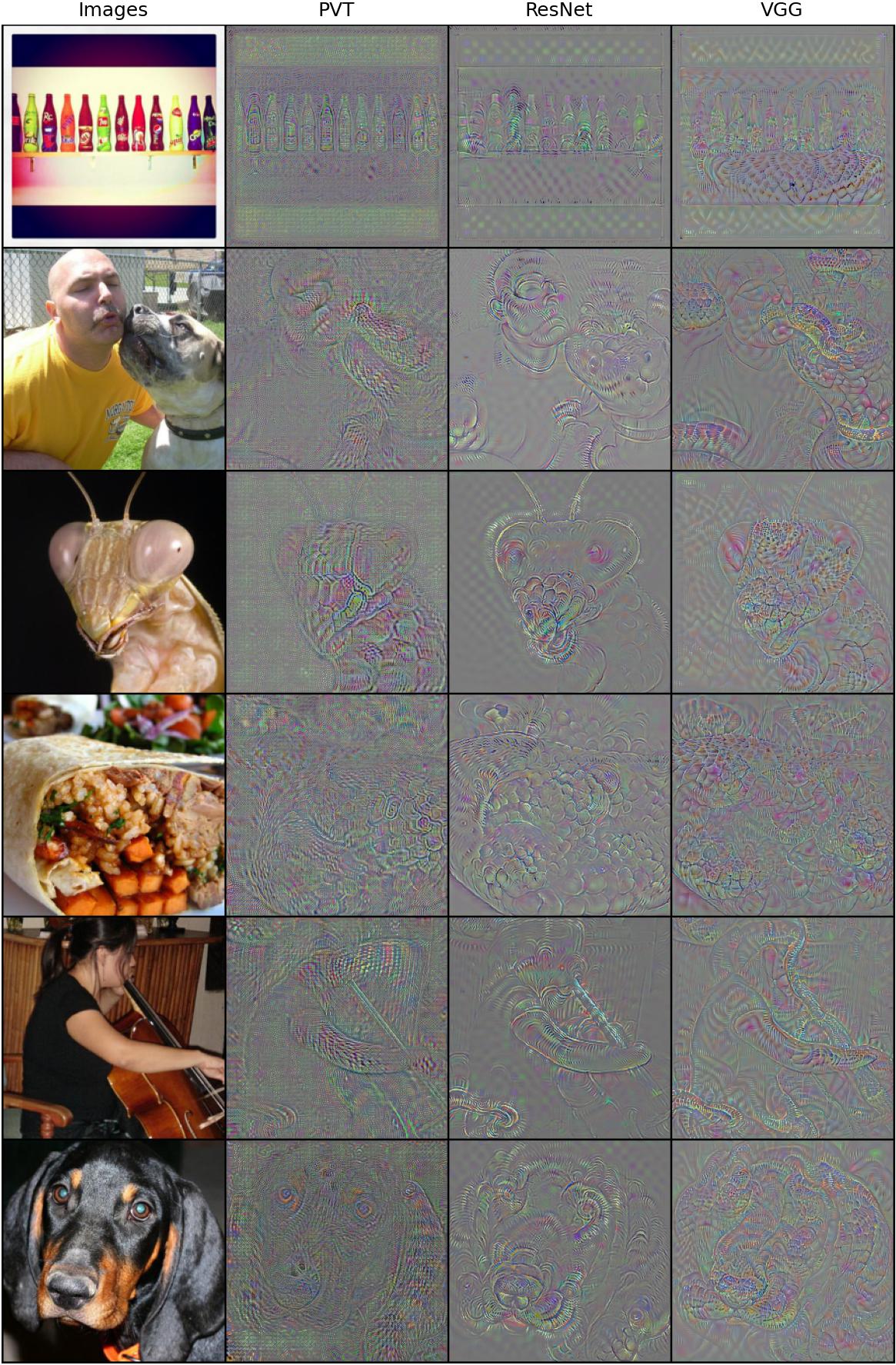}}
    \caption{Class accentuations (cf. Section~\ref{sec:accentuations}) for target class ``sidewinder'' for all studied models using 10-step PullbackAscent (cf. Section~\ref{sec:pullback_ascent}).}
    \label{fig:accentuations_sidewinder}
  \end{center}
\end{figure}

\begin{figure}[t]
\centering

\includegraphics[width=0.75\linewidth]
{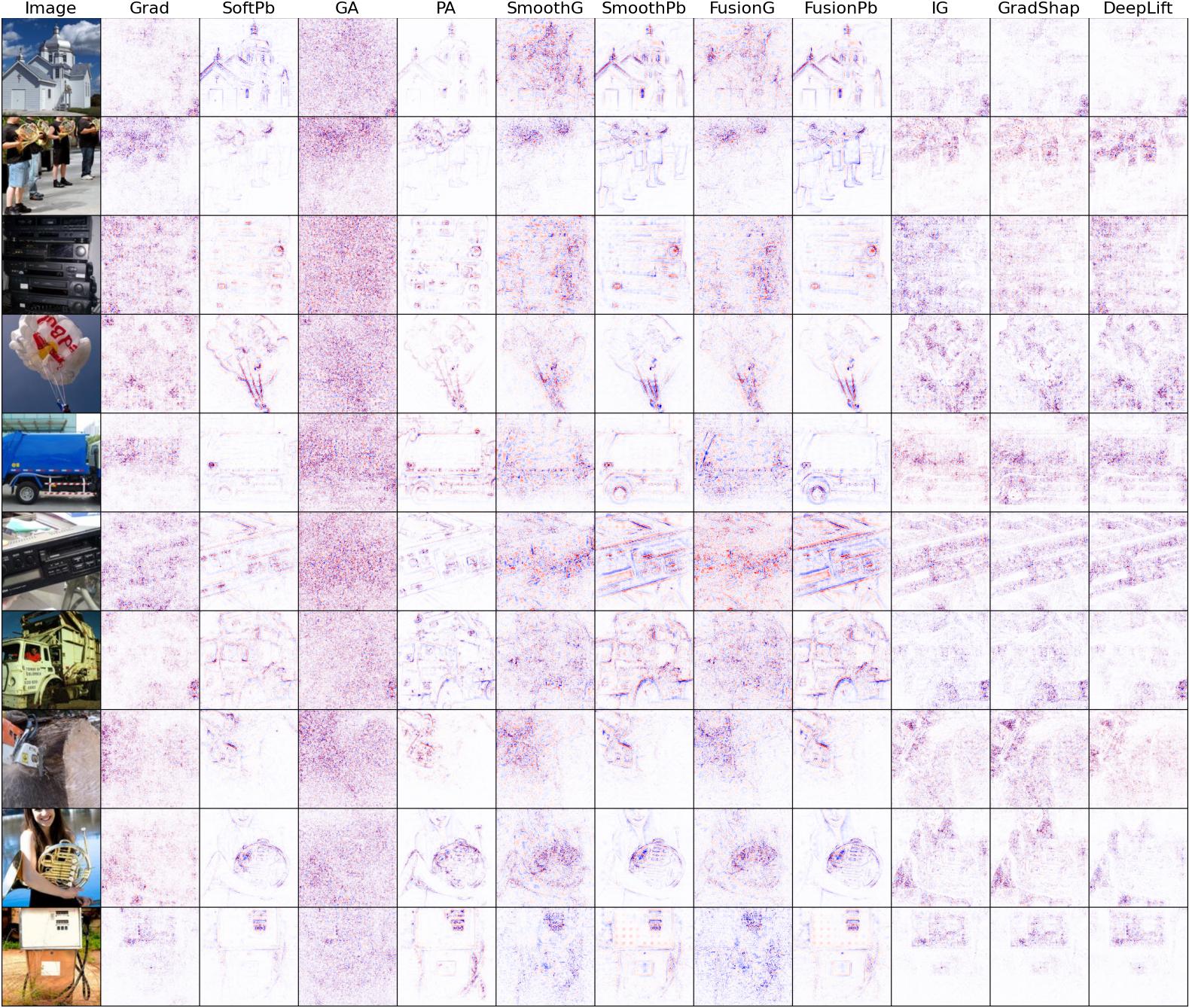}


\includegraphics[width=0.75\linewidth]
{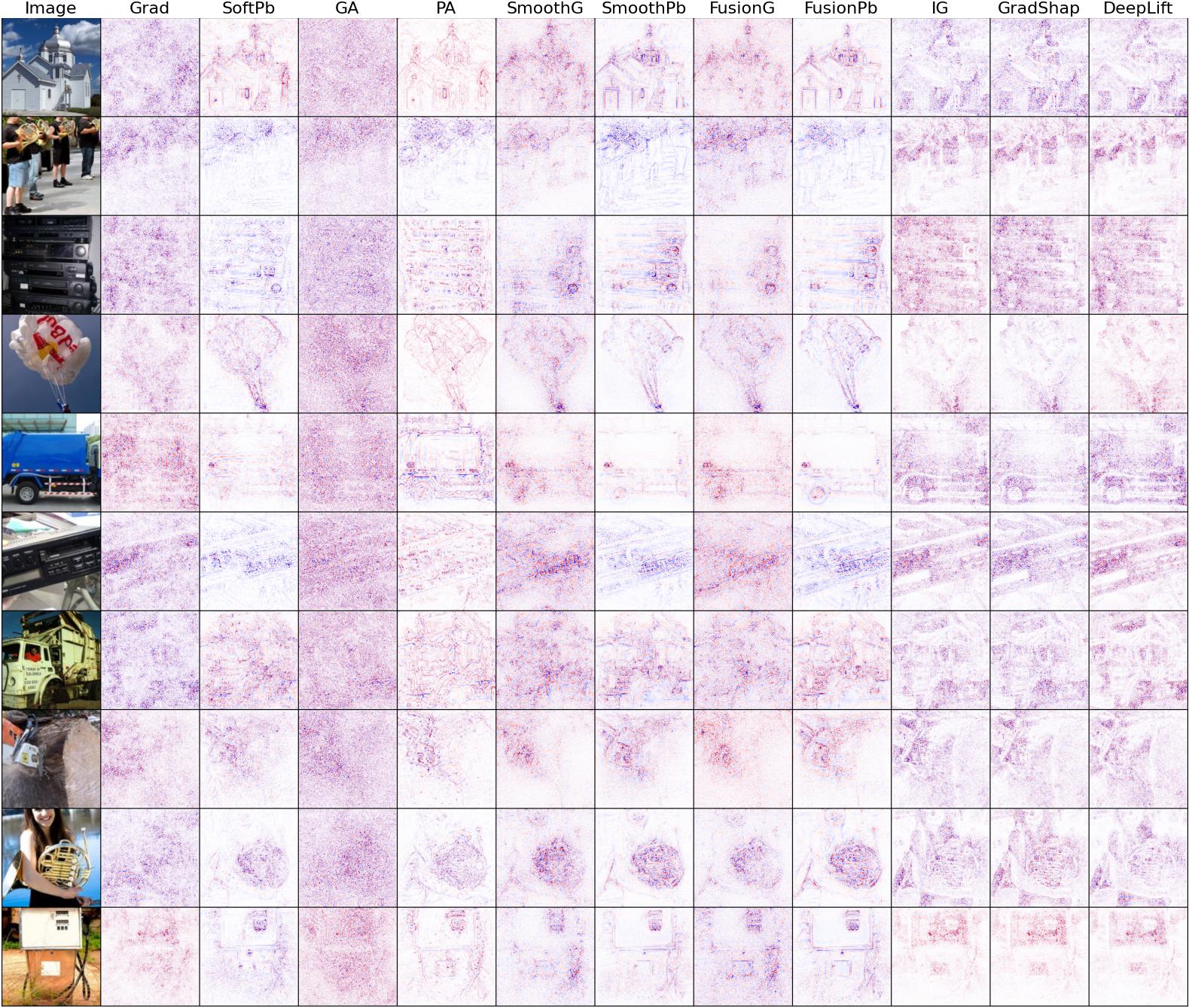}

\caption{
Qualitative comparison of explainers for ResNet50 (\textbf{top}) and VGG (\textbf{bottom}).
}
\label{fig:comparison_cnn}
\end{figure}

\begin{figure}[t]
\centering

\includegraphics[width=\linewidth]
{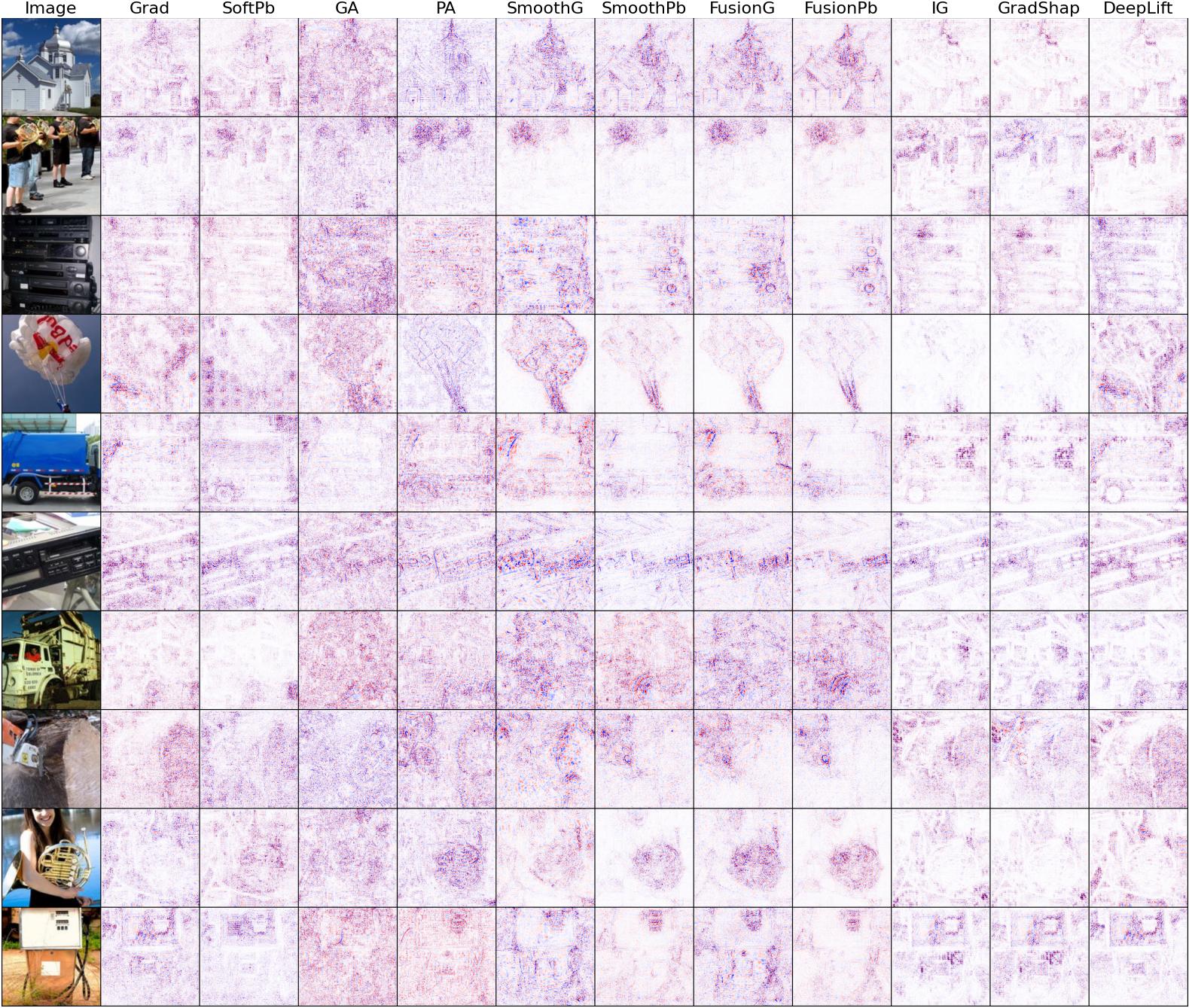}

\caption{
Qualitative comparison of explainers for PVT. Best viewed digitally.
}
\label{fig:comparison_pvt_v2_b1}
\end{figure}

This section provides additional experimental results, including runtime measurements (Table~\ref{table:time_complexity}), paired pullback-minus-gradient score differences with 95\% bootstrap confidence intervals (Table~\ref{table:paired_explainer_differences}), further examples of class-specific perturbations (Figures~\ref{fig:counterfactuals_vgg} and \ref{fig:counterfactuals_pvt}), target-specific feature accentuations (Figures~\ref{fig:accentuations_true_label}, \ref{fig:accentuations_mosque}, \ref{fig:accentuations_mud_turtle}, and \ref{fig:accentuations_sidewinder}), and qualitative comparisons of gradient- and pullback-based explanations (Figures~\ref{fig:comparison_cnn} and \ref{fig:comparison_pvt_v2_b1}).


\begin{table}[tb]
\setlength{\tabcolsep}{10pt}
\caption{Empirical attribution runtime of different explanation methods. Execution time is measured on a batch of 10 images on a local workstation with a 13th Gen Intel(R) Core(TM) i7-13620H CPU (2.40 GHz), 32GB RAM, and an NVIDIA GeForce RTX 4050 GPU with 6GB memory.}
\label{table:time_complexity}
\begin{center}
\begin{tabular}{lccc}
\toprule
Method
& ResNet50
& PVT-V2-B1
& VGG11 \\
\midrule
Gradient             & 0.064 s & 0.067 s & 0.074 s \\
\textbf{SoftPullback} & 0.097 s & 0.092 s & 0.130 s \\
\pairsep
GradientAscent        & 0.270 s & 0.289 s & 0.334 s \\
\textbf{PullbackAscent} & 0.419 s & 0.349 s & 0.577 s \\
\pairsep
SmoothGrad            & 2.632 s & 2.802 s & 3.349 s \\
\textbf{SmoothPullback} & 4.101 s & 3.222 s & 5.736 s \\
\pairsep
FusionGrad            & 5.713 s & 6.961 s & 12.185 s \\
\textbf{FusionPullback} & 8.819 s & 7.383 s & 12.587 s \\
\midrule
IntegratedGradients   & 9.448 s & 13.308 s & 13.754 s \\
GradientShap          & 0.423 s & 2.926 s & 1.083 s \\
DeepLift              & 0.222 s & 0.161 s & 0.575 s \\
\bottomrule
\end{tabular}
\end{center}
\end{table}



\begin{table}[tb]
\small
\caption{Pullback-minus-gradient score differences.
Each entry reports the mean paired difference and its 95\% percentile
bootstrap confidence interval, computed using 10,000 resamples.
Negative differences favour pullbacks for Infidelity and Max Sensitivity;
positive differences favour pullbacks for Focus, Faithfulness Correlation,
and Monotonicity Correlation.
Red entries indicate intervals lying entirely on the deterioration side
of zero; blue entries indicate intervals including zero, providing
inconclusive evidence of improvement or deterioration.
Intervals are not adjusted for multiple comparisons.
Each comparison uses 1000 paired observations.}
\label{table:paired_explainer_differences}
\begin{center}
\begin{tabular}{lccccc}
\toprule
Pair
& Infidelity $\downarrow$
& Focus $\uparrow$
& Faith.Corr $\uparrow$
& Mono.Corr $\uparrow$
& Max.Sens $\downarrow$ \\
\midrule
\multicolumn{6}{l}{\textbf{ResNet50}} \\
\addlinespace[2pt]
SoftPb - Grad
& \pairedci{-77.3050}{-81.3633}{-73.2595}
& \pairedci{0.1116}{0.1023}{0.1208}
& \pairedci{0.1472}{0.1226}{0.1710}
& \pairedci{0.1444}{0.1179}{0.1720}
& \pairedci{-0.8335}{-0.8431}{-0.8236} \\
PA - GA
& \pairedci{-24.0957}{-26.3064}{-21.9463}
& \pairedci{0.1419}{0.1353}{0.1487}
& \pairedci{0.1040}{0.0872}{0.1206}
& \pairedci{0.1903}{0.1656}{0.2153}
& \pairedci{-1.0230}{-1.0277}{-1.0181} \\
SmoothPb - SmoothG
& \pairedci{-58.7615}{-62.3481}{-55.1899}
& \pairedci{0.0733}{0.0664}{0.0802}
& \inconclusive{\pairedci{-0.0058}{-0.0218}{0.0100}}
& \inconclusive{\pairedci{-0.0128}{-0.0310}{0.0058}}
& \pairedci{-0.4187}{-0.4232}{-0.4142} \\
FusionPb - FusionG
& \pairedci{-48.3830}{-51.9694}{-44.9644}
& \pairedci{0.0572}{0.0509}{0.0637}
& \deterioration{\pairedci{-0.0385}{-0.0535}{-0.0231}}
& \inconclusive{\pairedci{-0.0073}{-0.0247}{0.0104}}
& \pairedci{-0.4932}{-0.4997}{-0.4866} \\
\midrule
\multicolumn{6}{l}{\textbf{PVT}} \\
\addlinespace[2pt]
SoftPb - Grad
& \pairedci{-2.6502}{-3.1834}{-2.1285}
& \pairedci{0.0599}{0.0541}{0.0655}
& \pairedci{0.0137}{0.0005}{0.0270}
& \deterioration{\pairedci{-0.0208}{-0.0390}{-0.0029}}
& \deterioration{\pairedci{0.0317}{0.0195}{0.0434}} \\
PA - GA
& \pairedci{-2.8716}{-3.1190}{-2.6317}
& \pairedci{0.0036}{0.0003}{0.0069}
& \inconclusive{\pairedci{0.0044}{-0.0109}{0.0200}}
& \pairedci{0.0544}{0.0362}{0.0731}
& \pairedci{-0.3875}{-0.3924}{-0.3824} \\
SmoothPb - SmoothG
& \pairedci{-3.8244}{-4.3282}{-3.3365}
& \pairedci{0.0411}{0.0363}{0.0459}
& \inconclusive{\pairedci{0.0059}{-0.0049}{0.0165}}
& \deterioration{\pairedci{-0.0483}{-0.0644}{-0.0325}}
& \pairedci{-0.0497}{-0.0540}{-0.0454} \\
FusionPb - FusionG
& \pairedci{-1.5167}{-1.9578}{-1.0738}
& \pairedci{0.0084}{0.0045}{0.0123}
& \inconclusive{\pairedci{-0.0088}{-0.0178}{0.0004}}
& \deterioration{\pairedci{-0.0617}{-0.0769}{-0.0462}}
& \pairedci{-0.2258}{-0.2363}{-0.2149} \\
\midrule
\multicolumn{6}{l}{\textbf{VGG}} \\
\addlinespace[2pt]
SoftPb - Grad
& \pairedci{-83.7863}{-90.4122}{-77.2206}
& \pairedci{0.0702}{0.0636}{0.0768}
& \pairedci{0.1489}{0.1319}{0.1657}
& \pairedci{0.0589}{0.0364}{0.0814}
& \pairedci{-0.6860}{-0.6919}{-0.6799} \\
PA - GA
& \pairedci{-12.9064}{-15.1887}{-10.8017}
& \pairedci{0.0666}{0.0619}{0.0712}
& \pairedci{0.0618}{0.0510}{0.0734}
& \pairedci{0.0778}{0.0562}{0.0987}
& \pairedci{-0.8275}{-0.8319}{-0.8230} \\
SmoothPb - SmoothG
& \pairedci{-38.0143}{-42.2369}{-33.8445}
& \pairedci{0.0630}{0.0588}{0.0674}
& \pairedci{0.0322}{0.0213}{0.0430}
& \deterioration{\pairedci{-0.0752}{-0.0904}{-0.0600}}
& \pairedci{-0.3831}{-0.3867}{-0.3794} \\
FusionPb - FusionG
& \pairedci{-22.5950}{-26.2341}{-19.3095}
& \pairedci{0.0530}{0.0490}{0.0570}
& \inconclusive{\pairedci{0.0070}{-0.0029}{0.0169}}
& \deterioration{\pairedci{-0.0483}{-0.0630}{-0.0342}}
& \pairedci{-0.4421}{-0.4478}{-0.4362} \\
\bottomrule
\end{tabular}
\end{center}
\end{table}


\end{document}